\documentclass[10pt,journal,compsoc]{IEEEtran}
\usepackage{amsmath,amsfonts}
\usepackage{array}

\usepackage{textcomp}
\usepackage{stfloats}
\usepackage{url}
\usepackage{verbatim}
\usepackage{longtable}
\usepackage{multirow}
\usepackage{graphicx}
\def\BibTeX{{\rm B\kern-.05em{\sc i\kern-.025em b}\kern-.08em
    T\kern-.1667em\lower.7ex\hbox{E}\kern-.125emX}}
\usepackage{balance}
\usepackage{amsmath}
\usepackage{amssymb}
\usepackage{subfigure}
\usepackage{color}
\usepackage[ruled]{algorithm2e}
\usepackage{float}
\usepackage[colorlinks,linkcolor=black]{hyperref}

\usepackage{pifont}
\usepackage[T1]{fontenc}
\usepackage[utf8]{inputenc}
\usepackage[switch]{lineno}

\usepackage[outdir=./]{epstopdf}
\usepackage{booktabs}
\usepackage{rotating}
\usepackage{tablefootnote}

\usepackage{algorithm}  
\usepackage{algpseudocode}  
\usepackage{amsmath}  
\renewcommand{\algorithmicrequire}{\textbf{Input:}}  
\renewcommand{\algorithmicensure}{\textbf{Output:}} 

\begin{document}

\title{Fuzzy Granule Density-Based Outlier Detection\\ with Multi-Scale Granular Balls}
\author{Can Gao \IEEEmembership{Member, IEEE}, Xiaofeng Tan, Jie Zhou, Weiping Ding, \IEEEmembership{Senior Member, IEEE}, \\and Witold Pedrycz \IEEEmembership{Life Fellow, IEEE}
\thanks{
\indent Can Gao and Xiaofeng Tan are with the College of Computer Science and Software Engineering, Shenzhen University, Shenzhen, 518060, China, and also with Guangdong Key Laboratory of Intelligent Information Processing, Shenzhen, 518060, China. (e-mail: 2005gaocan@163.com, txf0620@gmail.com)

Jie Zhou is with the National Engineering Laboratory for Big Data System Computing Technology, Shenzhen University, Shenzhen, 518060, China. (e-mail: jie\_jpu@163.com)

Weiping Ding is with the School of Artificial Intelligence and Computer Science, Nantong University, Nantong, 226019, China, and also the Faculty of Data Science, City University of Macau, Macau 999078, China. (e-mail: dwp9988@163.com)

Witold Pedrycz is with the Department of Electrical and Computer Engineering, University of Alberta, Edmonton, Canada, with Systems Research Institute, Polish Academy of Sciences, Warsaw, Poland, with the Department of Electrical and Computer Engineering, King Abdulaziz University, Jeddah, Saudi Arabia, and also with the Department of Computer Engineering, Istinye University, Istanbul, Turkey. (e-mail: wpedrycz@ualberta.ca).

Manuscript received xx, xxxx; revised xx, xxxx. (Corresponding author: Jie Zhou)
}}

\markboth{IEEE TRANSACTIONS ON KNOWLEDGE AND DATA ENGINEERING, ~Vol.~xx, No.~xx, December~2024}%
{How to Use the IEEEtran \LaTeX \ Templates}

\IEEEtitleabstractindextext{%
\begin{abstract}
Outlier detection refers to the identification of anomalous samples that deviate significantly from the distribution of normal data and has been extensively studied and used in a variety of practical tasks. However, most unsupervised outlier detection methods are carefully designed to detect specified outliers, while real-world data may be entangled with different types of outliers. In this study, we propose a fuzzy rough sets-based multi-scale outlier detection method to identify various types of outliers. Specifically, a novel fuzzy rough sets-based method that integrates relative fuzzy granule density is first introduced to improve the capability of detecting local outliers. Then, a multi-scale view generation method based on granular-ball computing is proposed to collaboratively identify group outliers at different levels of granularity. Moreover, reliable outliers and inliers determined by the three-way decision are used to train a weighted support vector machine to further improve the performance of outlier detection. The proposed method innovatively transforms unsupervised outlier detection into a semi-supervised classification problem and for the first time explores the fuzzy rough sets-based outlier detection from the perspective of multi-scale granular balls, allowing for high adaptability to different types of outliers. Extensive experiments carried out on both artificial and UCI datasets demonstrate that the proposed outlier detection method significantly outperforms the state-of-the-art methods, improving the results by at least 8.48\% in terms of the Area Under the ROC Curve (AUROC) index. { The source codes are released at \url{https://github.com/Xiaofeng-Tan/MGBOD}. }

\end{abstract}

}

\maketitle

\IEEEdisplaynontitleabstractindextext

\IEEEpeerreviewmaketitle

\section{Introduction}
\IEEEPARstart{D}{etecting} samples or patterns that deviate from the majority of data are both meaningful and practical, and helps to monitor risky behaviors, rare events, and abnormal observations in real-world tasks. Outlier detection \cite{Boukerche2021, Ma2021, Pang2022}, also known as anomaly detection, is an effective method that aims to find anomalous samples from the data. Due to its practicality and effectiveness, outlier detection has become an important research topic in the fields of data mining and machine learning, and has been extensively used in a variety of domains such as intrusion detection \cite{WangYZ2023}, financial fraud \cite{Chen2022}, and video surveillance \cite{Zhou2019}. 

Existing outlier detection methods \cite{Boukerche2021} can be categorized into supervised, semi-supervised, and unsupervised methods in terms of the availability of labeled data. Among them, unsupervised methods have attracted increasing attention and play a crucial role in outlier detection. Generally speaking, unsupervised outlier detection methods \cite{Boukerche2021} can be divided into : statistical-based methods \cite{Reynolds2009}, distance-based methods \cite{Angiulli2002}, density-based methods \cite{Breunig2000}, clustering-based methods \cite{He2003}, ensemble-based methods \cite{Lazarevic2005}, and others \cite{Ramaswamy2000}. Owing to the lack of label information, these unsupervised outlier detection methods have been carefully designed with specific assumptions. For example, statistic-based methods generally assume that normal samples follow a prior distribution, allowing the identification of outliers by using statistical information. Distance-based methods suppose that outliers are far from inliers, and thus the distance between a sample to its neighbors can be used as an effective measure of outlierness. Density-based methods postulate that normal samples are aggregated in regions with high density, and a sample with low density can be considered an outlier. While clustering-based methods are performed with the assumption that samples can be grouped into clusters, and outliers are either difficult to cluster or located in small clusters. Obviously, the performance of these methods highly relies on the conformity of the assumptions with underlying outliers. Moreover, most of these methods are developed to deal with numerical data and may be ineffective in detecting outliers from data with categorical attributes.

Rough set theory (RST) \cite{Pawlak1982, XiaBW2023}, as an effective soft computing method for knowledge representation and approximate reasoning, is effective in handling categorical data entangled with vagueness, imprecision, and uncertainty. Rough sets have also been used for outlier detection because of the superiority of representing the relationship of samples with categorical attributes \cite{Jiang2009}. Jiang et al. \cite{Jiang2008} explored the theory of rough sets for identifying outliers and proposed a rough membership function-based outlier detection method. Subsequently, Jiang et al. \cite{Jiang2011} introduced a hybrid outlier detection method that integrates distance information with the boundary region in rough sets. Albanese et al. \cite{Albanese2014} studied the problem of outlier detection in spatiotemporal data and presented a rough outlier set extraction method based on the lower and upper approximations in rough sets. From the perspective of information theory, Jiang et al. \cite{Jiang2010} defined the information entropy-based outliers and developed an effective algorithm to find these outliers. Further, Jiang et al. \cite{Jiang2019} combined approximation accuracy with conditional entropy and presented an approximation accuracy entropy-based outlier detection method. In addition, Maciá-Pérez et al. \cite{F2015} proposed a rough sets-based computationally efficient method for detecting outliers from large-scale data.

Nevertheless, the above rough sets-based methods rely on the equivalence relation to form sample granules, making it difficult to handle data with numerical or mixed attributes. To address this problem, fuzzy rough sets (FRS) \cite{Hu2010} and neighborhood rough sets (NRS) \cite{Xia2023a} have been introduced for outlier detection. Chen et al. \cite{Chen2010} introduced the concept of neighborhood-based outlier factor and proposed the neighborhood rough sets-based outlier detection method. Yuan et al. \cite{Yuan2018, Yuan2022a} used neighborhood information entropy and multigranulation relative entropy to compute the outlier factor of samples and developed the corresponding outlier detection algorithms for mixed data. Gao et al. \cite{Gao2023} defined a ratio and negative region detection factor and provided a relative granular ratio-based outlier detection method. Zhang et al. \cite{ZhangYM2023} analyzed the neighborhood structure in neighborhood rough sets using the three-way decision and presented a multiple neighborhood outlier factor-based outlier detection method for heterogeneous data. By introducing fuzzy similarity relation, Yuan et al. have proposed several outlier detection methods based on the concepts of fuzzy information entropy \cite{Yuan2021}, fuzzy rough granule \cite{Yuan2022}, multi-fuzzy granules \cite{Yuan2023a}, and fuzzy rough density \cite{Yuan2023b}, respectively, and achieved promising results on both numerical and categorical data.

Most aforementioned unsupervised outlier detection methods have been carefully designed to identify a certain type of outliers. However, practical data may contain different types of outliers such as local outliers, global outliers, and group outliers \cite{Han2022}, and these methods may face challenges in dealing with such scenarios. Moreover, some outliers such as group outliers may be difficult to identify in a single view, and exploring multi-view information to detect these outliers may be a promising way to address this problem \cite{Tan2023}. In this study, we propose an improved fuzzy rough sets-based multi-scale outlier detection method to identify different types of outliers. To sum up, the contribution of this paper is threefold.
\begin{enumerate}
    \item [(1)] To accurately identify local outliers, a novel fuzzy rough sets-based outlier detection method is proposed, which integrates sample fuzzy similarity with relative fuzzy granule density and significantly enhances the separability of local outliers. 
    \item [(2)] To effectively detect group outliers, a multi-scale view generation method based on granular-ball computing is introduced, which enables multi-view outlier detection and provides the capability of identifying outliers at different levels of granularity.
    \item [(3)] To further improve the outlier detection results, a weighted support vector machine is developed, which is trained on reliable outliers and inliers determined by the three-way decision and facilitates the detection of outliers from uncertain samples. Extensive experiments and statistical significance analysis demonstrate that the proposed method can effectively detect different types of outliers, outperforming other state-of-the-art methods by a large margin.
\end{enumerate}


The remaining sections of the paper are organized as follows. Section \ref{section 2} presents preliminaries on fuzzy rough sets and granular ball-based learning. Section \ref{section 3} elaborates on the proposed multi-scale outlier detection method. The experimental results and statistical significance analysis are reported in Section \ref{section 4}. Section \ref{section 5} concludes the paper and offers some feature research directions.

\section{Preliminaries} \label{section 2}
This section presents some related concepts in fuzzy rough sets and granular ball-based learning, whose detailed information can be referred to \cite{Dubois1990, Xia2022a}.

\subsection{Fuzzy rough sets}
In fuzzy rough sets, the available data consist of a set of samples $U$ described by a finite attribute set $A$. On each attribute $a \in A$, a mapping function $f$ associates each sample $x \in U$ with a value from the domain of the attribute $a$, i.e., $f(x, a) \in V_a$. If there is no decision attribute in $A$, the data is also referred as to a fuzzy information system and formally represented as $FIS=(U, A, V, f)$, with $V=\bigcup_{a\in A} V_a$.

\noindent \textbf{Definition 1} (fuzzy relation) A fuzzy relation $R$ on $U$ is defined as $R: U\times U  \mapsto [0, 1]$ and satisfies the following properties: \cite{Dubois1990}
\begin{enumerate}
    \item [(1)]	Reflexivity: $R(x,x)=1$;
    \item [(2)]	Symmetry: $R(x,y)=R(y,x)$;
    \item [(3)]	Transitivity: $R(x,z) \geq \vee_{y\in U} \big(R(x,y)\wedge R(y,z) \big)$,
\end{enumerate}
where $R(x, y)$ denotes the degree of fuzzy similarity between the samples $x$ and $y$ given the fuzzy relation $R$. 

\noindent \textbf{Definition 2} (fuzzy granule) Let $FIS=(U, A, V, f)$ be a fuzzy information system. For any attribute subset $B\subseteq A$, it determines a fuzzy relation $R_B$ that forms a family of fuzzy granules on $U$, which is defined as \cite{Dubois1990}
\begin{equation}
    \begin{aligned}
        U/R_B=\{[x_1]_{B},[x_2]_{B},\ldots,[x_i]_{B},\ldots,[x_n]_{B}\},
    \end{aligned}
\end{equation}
where $[x_i]_{B}$ denotes the fuzzy granule of the sample $x_i$, which is defined as $[x_i]_{B} = \big(R_B(x_i, x_1),  R_B(x_i, x_2), \linebreak \ldots, R_B(x_i, x_n) \big)$, and the cardinality of the fuzzy granule $[x_i]_{B}$ is calculated as $|[x_i]_{B}|=\sum_{j=1}^{n}{R_B(x_i,x_j)}$, with $1\le|[x_i]_{B}|\le n$. 

\noindent \textbf{Definition 3} (fuzzy relation matrix) Let $R_B$ be the fuzzy relation induced by an attribute subset $B\subseteq A$. The fuzzy relation of samples given $R_B$ is denoted as \cite{Dubois1990}
\begin{equation}
    \begin{aligned}
        M(R_B)= 
        \begin{bmatrix}
            R_B(x_1\, x_1)&\cdots&R_B(x_1, x_n)\\
            \vdots&\ddots&\vdots\\
            R_B(x_n, x_1)&\cdots&R_B (x_n, x_n)\\
        \end{bmatrix},   
    \end{aligned}
\end{equation}
where $R_B(x_i, x_j)$ is the degree of fuzzy similarity between the samples $x_i$ and $x_j$ with respect to $B$. In this study, it can be calculated by $R_B(x_i,x_j)=\bigwedge_{a\in B}{{R_a}(x_i, x_j)}$, where $R_a(x_i, x_j)$ is defined as \cite{Yuan2021}
\begin{equation}\label{e3}
    \begin{aligned}
    & R_{a} (x_i,x_j) =\\&
        \begin{cases}
            1, & f_a(x_i)=f_a(x_j)\\ 
             & {\rm and\;}a{\rm \;is\;nominal,}\\
            0, & f_a(x_i)\neq f_a(x_j)\\
             &{\rm and\;}a{\rm \;is\;nominal,}\\
            1-|f_a(x_i)-f_a(x_j)|, &|f_a(x_i)-f_a(x_j)|\le\epsilon_a\\
            & {\rm \;and\;}a{\rm \ is\ numerical},\\
            0, &|f_a(x_i)-f_a(x_j)| > \epsilon_a\\
            & {\rm \;and\;}a{\rm \ is\ numerical},
        \end{cases}
    \end{aligned}
\end{equation}
where $\epsilon_a$ is a threshold parameter and calculated by $\epsilon_a=\frac{std(a)}{\delta}$, and $std(a)$ is the standard deviation of attribute values on $a$, and $\delta$ is an adjustable parameter. 

For nominal attributes, the degree of fuzzy similarity between the samples is equal to 1 when their values are identical; otherwise, their fuzzy similarity is 0. For numerical attributes, if the absolute difference in their values is lower than the threshold parameter, their fuzzy similarity is determined by their absolute difference; otherwise, they are considered completely dissimilar.

\noindent \textbf{Definition 4} (fuzzy upper and lower approximations) Let $FIS=(U, A, V, f)$ be a fuzzy information system and $B$ be an attribute subset of $A$. For a sample subset $X \subseteq U$, the fuzzy upper and lower approximations of a sample $x$ to $X$ with respect to $B$ are defined as \cite{Dubois1990}
\begin{equation}
    \begin{aligned}
        \begin{cases}
        \overline{R_B}X={\rm sup}_{y\in U} \mathcal{T} \{R_B(x,y),X(y)\},\\
        \underline{R_B}X={\rm inf}_{y\in U} \mathcal{S} \{N(R_B(x,y)),X(y)\},
        \end{cases}
    \end{aligned}
\end{equation}
where $\mathcal{T}$ and $\mathcal{S}$ denote the fuzzy triangular norm ($\mathcal{T}$-norm) and conorm ($\mathcal{S}$-norm), respectively, and $N$ is a negator, i.e., $N(x)=1-x$. In this study, the standard min and max fuzzy operators are used, i.e., $\mathcal{T} = \min$ and $\mathcal{S} = \max$.

\subsection{Granular ball-based learning}
Granular computing \cite{YaoVP2013} is known as a philosophy and methodology that simulates human thinking to solve complex problems, which encompasses all the theories, methods, and techniques related to information granularity. It solves complex problems by granulating and abstracting them into simpler sub-problems under different levels of granularity and has been widely applied in fields such as intelligent information processing, decision-making with uncertainty, and knowledge discovery. Inspired by the cognitive mechanism of the human brain, Xia, et al \cite{Xia2019} proposed the method of granular ball computing, which represents data with granular balls to perform clustering and classification tasks.

\noindent \textbf{Definition 5} (granular ball) Let $U$ be the set of samples in a fuzzy information system $FIS=(U, A, V, f)$ and $GB$ be the granular ball that contains a set of similar samples from $U$. The set of granular balls $GBS=\{GB_1, GB_2,\ldots, GB_n\}$ that are used to describe the sample set $U$ can be optimized by \cite{Xia2022a}
\begin{equation}
    \begin{aligned}
        &\min\;{\lambda_1}|U|/\sum_{GB_i\in GB}|GB_i|+\lambda_2|GB|,\\
        &{\rm s.t.}\; {quality}(GB_i)\geq T,
        \end{aligned}
        \label{e6}
\end{equation}
where the symbol ``$|\cdot|$'' denotes the cardinality of a sample set, $\lambda_1$ and $\lambda_2$ are the corresponding weight parameters, $quality(\cdot)$ and $T$ represent the measure for the quality of granular balls and the parameter for controlling the quality, respectively.

The first term of the objective function reflects the coverage of samples by granular balls and the second term indicates the number of generated granular balls. A higher coverage of granular balls results in less sample information loss, while a larger number of granular balls leads to a more accurate characterization of data but makes the learned model more complicated. As a result, the optimization objective of granular balls is to maximize the coverage samples by granular balls and minimize the number of granular balls, with the parameters to trade off the significance of different terms. For simplicity, following the work \cite{Xie2023}, these two parameters $\lambda_1$ and $\lambda_2$ are both set to 1 in this study. Also, the measure $quality(\cdot)$ and the parameter $T$ for the quality of granular balls use the default settings in the granular ball clustering method \cite{Xie2023}. 

Additionally, granular ball computing has been used in a wide variety of machine learning tasks, including granular ball classifiers \cite{Xia2019}, granular ball clustering \cite{Xie2023}, granular ball neighborhood rough set \cite{Xia2022b}, and granular ball sampling \cite{Xia2023}.

\section{The proposed method}\label{section 3}
In this section, we first outline the overall framework of the proposed multi-scale outlier detection method. Then, we elaborate on the improved fuzzy granule density-based outlier detection method as well as multi-scale granular ball computing. Finally, we present the multi-scale outlier detection method using a weighted support vector machine.

\subsection{Overall framework}
\begin{figure*}[!ht]
	\footnotesize
	\begin{centering}
		\includegraphics[scale=0.072]{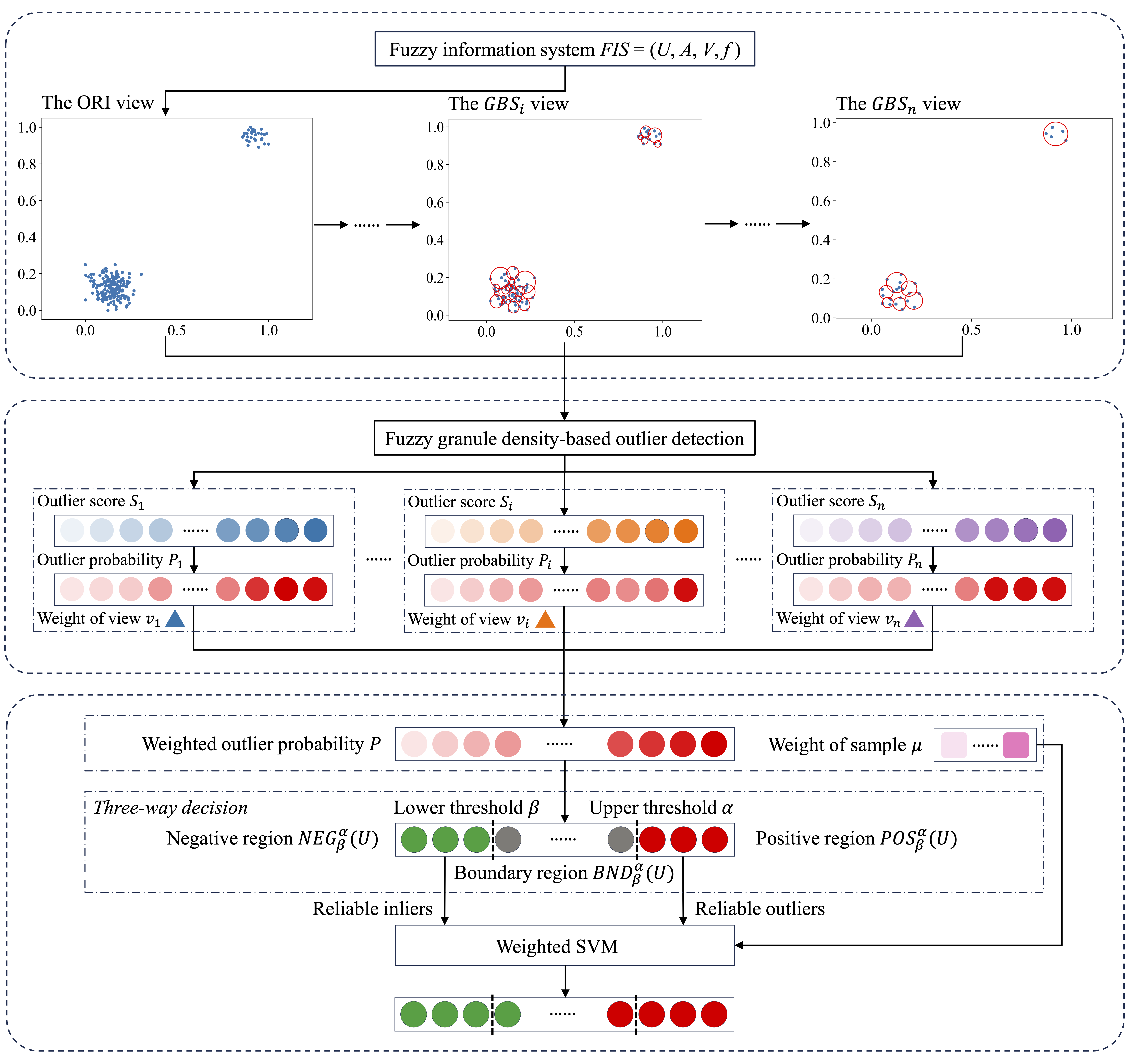}
		\par\end{centering}
	\caption{Framework of multi-scale outlier detection with granular balls.
        \label{f1}}
\end{figure*}

FRS is an effective method for outlier detection, which utilizes the fuzzy similarity relation to granulate samples and further employs the concept of fuzzy granules to compute the outlier score of each sample. Intrinsically, FRS-based outlier detection is a distance-based method that excels at identifying global outliers. However, practical data may contain different types of outliers, such as local, global, and group outliers, which poses a substantial challenge to FRS-based outlier detection methods. In fact, density information plays a crucial role in identifying local outliers. By estimating the sample density in the local region, some outliers located in low-density regions can be detected efficiently. On the other hand, multi-scale views are inherently beneficial for detecting different types of outliers. For instance, in the higher scale view, group outliers are aggregated into easily recognizable samples, while in the lower scale view, more detailed information is provided, which helps to detect local and global outliers. Motivated by these facts, a multi-scale outlier detection method based on FRS and granular balls is proposed, and the overall framework is shown in Fig. \ref{f1}.

Specifically, multi-scale representations of the raw data are generated by using the technique of granular ball computing. In each scale, the improved FRS-based outlier detection method that integrates density information is used to identify outliers. To align the results obtained by different levels of scales, each set of outlier scores is first mapped into a probability vector, and then all samples are divided into positive, boundary, and negative regions according to their outlier probability fused from outlier probability vectors at different scales. By using reliable positive and negative samples, a weighted Support Vector Machine (SVM) is finally trained to determine the probability of each sample being an outlier.

\subsection{Fuzzy rough sets with relative granule density for outlier detection}
To overcome the drawback that RST can not efficiently handle numerical data, FRS has been introduced for outlier detection \cite{Yuan2022, Yuan2023a}. In fact, the FRS-based methods are good at detecting global outliers but ineffective in identifying local outliers since the fuzzy similarity relation is used to measure the distance of samples. On the contrary, density-based outlier detection methods are effective in detecting local outliers. Intuitively, introducing density information into FRS is beneficial to improve the performance of outlier detection. In this study, the relative density information is integrated into FRS-based outlier detection methods to recognize both global and local outliers.

In unsupervised scenarios, outlier detection methods aim to identify a set of anomalous samples from given data. For FRS-based outlier detection methods, the objective function can be formalized as
\begin{equation}
    \begin{aligned}
        &\min_O{\sum_{x_i\in O}\sum_{x_j\in U-O}{R_A(x_i,x_j)}},\\
        &{\rm s.t.}\;O\subset U,|O|=o,
        \end{aligned}\label{e7}
\end{equation}
where $O$ is the set of outliers, $A$ is the set of attributes in a fuzzy information system, the symbol $|\cdot|$ denotes the cardinality of a set, and $o$ is the number of outliers.

By using FRS-based outlier detection methods, outliers, particularly global outliers, could be identified effectively. However, existing methods encounter challenges in detecting local outliers, which exhibit similar characteristics to inliers but differ in density \cite{Breunig2000}. Considering both the distance and density information, the objective function can be expressed as
    \begin{equation}
        \begin{aligned}
                & \min_{O_g,O_l}\sum_{x_i\in O_g}\sum_{x_j\in U-O_g}{R_A(x_i,x_j)} \\
                & +\sum_{x_i\in O_l}\sum_{x_j\in U-O_l} exp\{-\lambda \Vert Den_A(x_i) - Den_A(x_j) \Vert_2^2 \},\\
                & {\rm s.t.}\; O_l,O_g\subset U, O_l\cap O_g = \varnothing, |O_g|+|O_l|=o,
        \end{aligned}
        \label{e8}
    \end{equation}
where $O_g$ and $O_l$ denote the set of global and local outliers, respectively, $Den_A(x)$ denotes the density of the sample $x$ with respect to the attribute set $A$, $\lambda$ is an adjustable parameter to control the strength of density information, and $exp$ stands for the exponential function.

In real-world applications, there is no clear difference between local and global outliers, making it challenging to obtain the optimal solution to the problem (\ref{e8}). For brevity, the objective function is approximately expressed as
    \begin{equation}
        \begin{aligned}
        & \min_O{\sum_{x_i\in O}\sum_{x_j\in U-O}{{(R}_A(x_i,x_j)}} \\
        & +exp\{-\lambda \Vert Den_A(x_i)-Den_A(x_j) \Vert_2^2 \}),\\
        & {\rm s.t.}\;O\subset U,|O|=o.
        \end{aligned}
        \label{e9}
    \end{equation}

It is observed that the problem (\ref{e9}) is not easy to optimize directly. Alternatively, an approximate objective function is presented as
\begin{equation}
    \begin{aligned}
    & \min_O{\sum_{x_i\in O}\sum_{x_j\in U-O}{{(R}_A(x_i,x_j)}} \\ \cdot 
    & exp\{-\lambda \Vert Den_A(x_i)-Den_A(x_j) \Vert_2^2 \}),\\
    & {\rm s.t.}\;O\subset U,|O|=o.
    \end{aligned}
    \label{e10}
\end{equation}

\begin{figure}[!ht]
	\footnotesize
	\begin{centering}
		\subfigure[]{\includegraphics[scale=0.035]{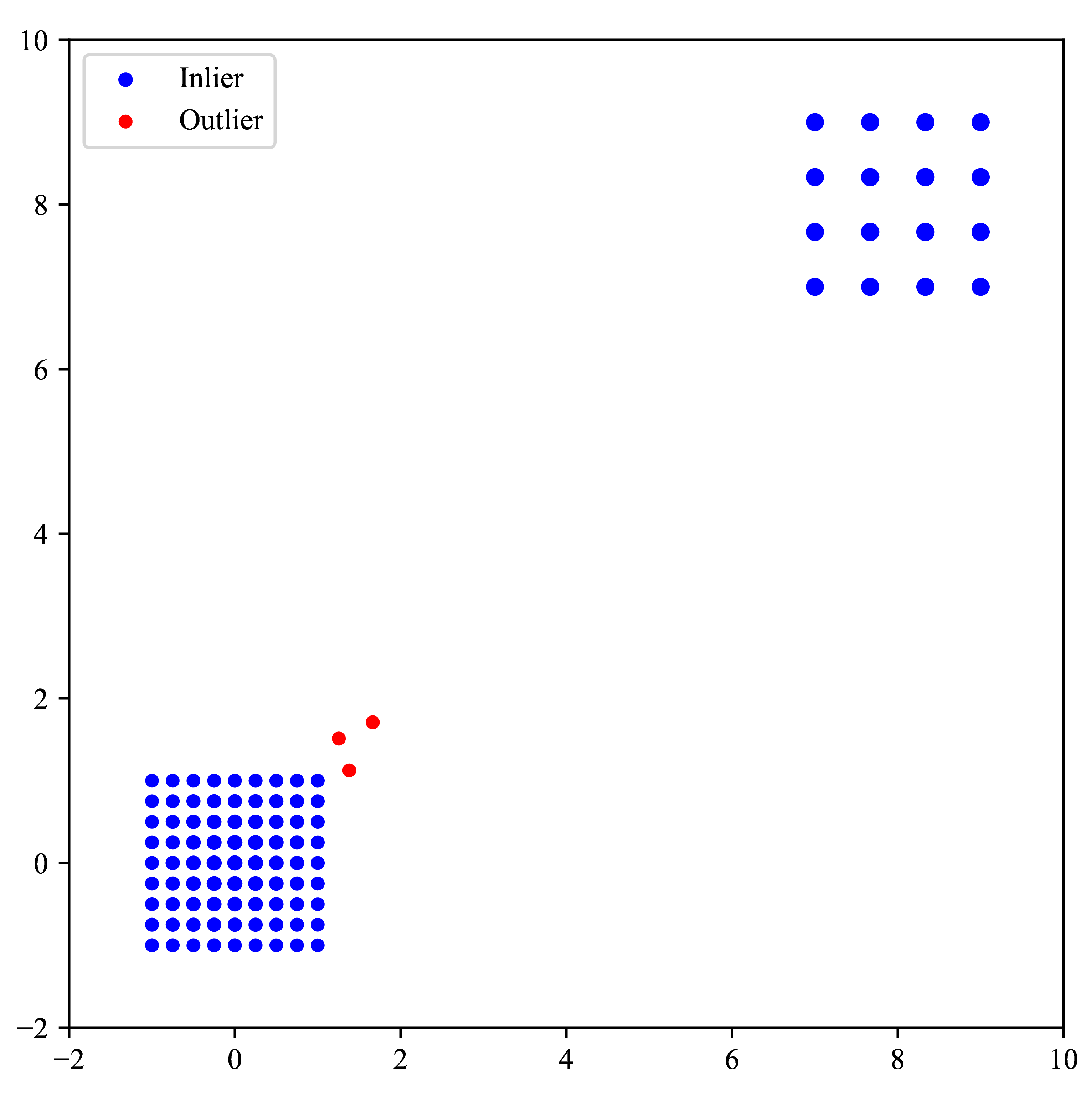}} \label{f2-a}
		\subfigure[]{\includegraphics[scale=0.045]{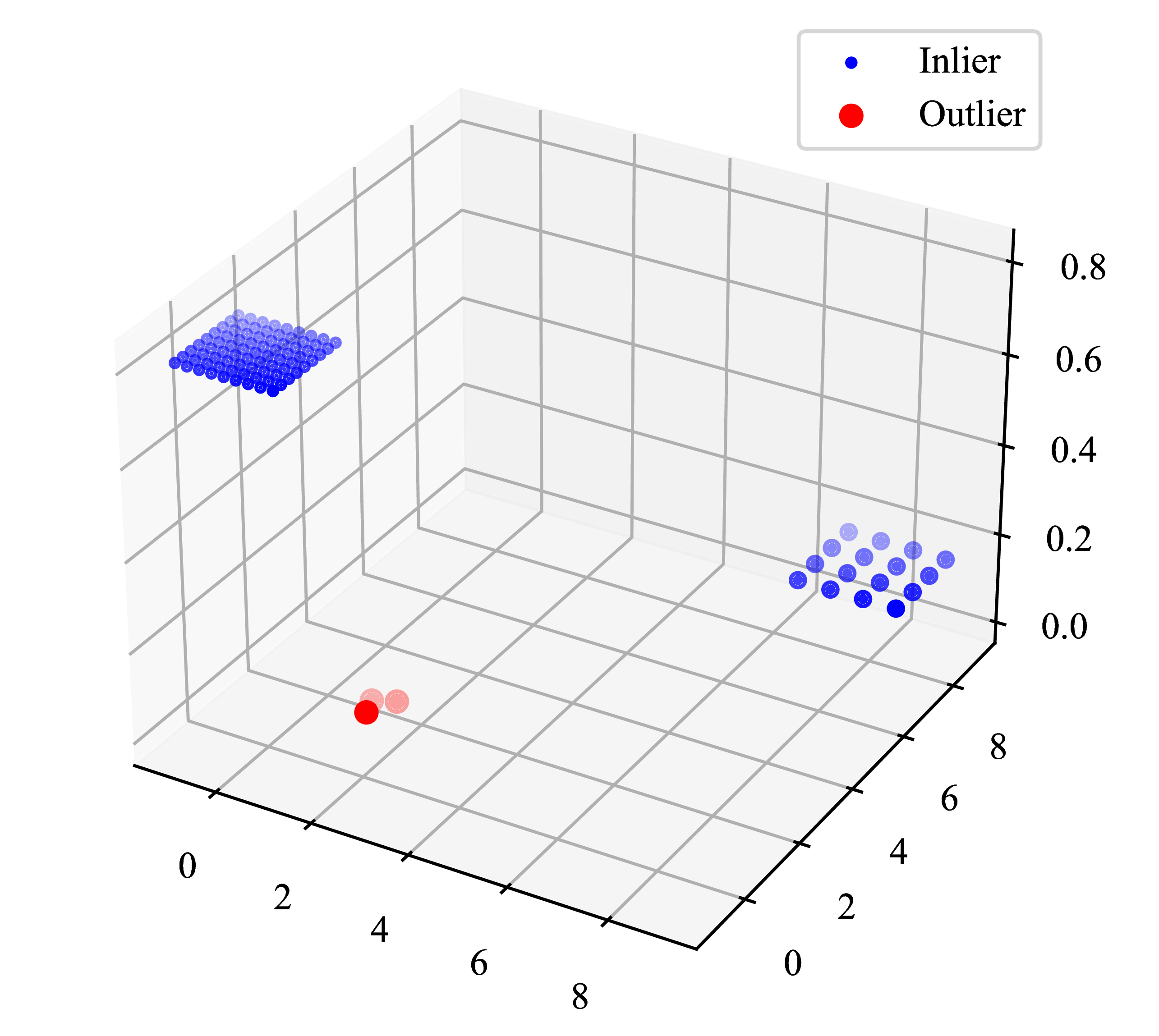}}  \label{f2-b}
	 	\par\end{centering}
	\caption{The results of local outlier detection on the synthetic dataset. a) The results of outlier detection without density information; b) The results of outlier detection using density information (visualized in 3-dimensional space).}\label{f2}
\end{figure}

After transforming the objective function, the separability between local outliers and normal samples is intrinsically enhanced. Fig. \ref{f2} shows a synthetic dataset that consists of some inliers with different densities and local outliers, which are marked in blue and red, respectively, with the size to reflect the magnitude of being an outlier. As shown in Fig. \ref{f2} (a), it is obvious that samples located in the upper right corner are more likely to be identified as outliers when using distance-based outlier detection methods, like FRS-based methods. But in Fig. \ref{f2} (b), it is observed that local outliers are well separated from inliers after introducing density information, and consequently, FRS-based outlier detection methods can achieve better performance in this scenario. Motivated by this fact, we propose a novel FRS-based outlier detection method that integrates with density information. In what follows, we introduce some relevant concepts of the proposed method.

\noindent \textbf{Definition 6} (fuzzy granule density) For any sample $x \in U$, its fuzzy granule density with respect to an attribute $a \in A$ is defined as \cite{Yuan2023a}
\begin{equation}
    \begin{aligned}
    {Den}_a(x)=\frac{|[x]_a|}{|U|}, 
    \end{aligned}
\end{equation}
where $|[x]_a|$ denotes the cardinality of the fuzzy granule $[x]_a$.

\noindent \textbf{Definition 7} (relative fuzzy granule density) For any two samples $x_i, x_j \in U$, their relative fuzzy granule density with respect to an attribute $a \in A$ is defined as
\begin{equation}
    \begin{aligned}
    {Rel\_Den}_a(x_i,x_j)=exp\{-\lambda \Vert Den_a(x_i)-Den_a(x_j) \Vert_2^2\},
    \end{aligned}
\end{equation}
where $\lambda$ is a weighted parameter, and $\Vert \cdot \Vert_2$ denotes 2-norm.

\noindent \textbf{Definition 8} (fuzzy similarity degree with density) For any two samples $x_i, x_j \in U$, their fuzzy similarity degree using density with respect to an attribute $a \in A$ is defined as
\begin{equation}
    \begin{aligned}
    \widetilde{R}_a(x_i,x_j)=R_a(x_i,x_j){Rel\_Den}_a(x_i,x_j).
    \end{aligned}
\end{equation}

\noindent \textbf{Definition 9} (fuzzy similarity relation with density) For any attribute subset $B\subseteq A$, the fuzzy relation with density $\widetilde{R}_B$ is defined as
\begin{equation}
    \begin{aligned}
        U/\widetilde{R}_B=\{\widetilde{[x_1]}_{B}, \widetilde{[x_2]}_{B},\ldots,\widetilde{[x_i]}_{B},\ldots, \widetilde{[x_n]}_{B}\},
        \end{aligned}
\end{equation}
where $\widetilde{[x_i]}_{B}$ denotes the fuzzy granule with density of the sample $x_i$, which is defined as $\widetilde{[x_i]}_{B}=\big(\widetilde{R}_B(x_i,x_1), \widetilde{R}_B(x_i, \linebreak x_2),  \ldots, \widetilde{R}_B(x_i,x_n)\big)$ and $\widetilde{R}_B(x_i,x_j)=\bigwedge_{a \in B}{\widetilde{R}_{a}(x_i,x_j)}$, and the cardinality of the fuzzy granule with density $\widetilde{[x_i]}_{B}$ is calculated as $|\widetilde{[x_i]}_{B}|=\sum_{j=1}^{n}{\widetilde{R}_B(x_i,x_j)}$, with $1\le|\widetilde{[x_i]}_{B}|\le n$. 

\noindent \textbf{Definition 10} (significance) For any attribute subset $B \subseteq A$, the significance of $B$ is defined as
\begin{equation}
    \begin{aligned}
    Sig(B)=-\log \sum_{x\in U}\frac{|\widetilde{[x]}_{B}|}{|U|},
    \end{aligned}
\end{equation}

\noindent \textbf{Definition 11} (attribute sequence) Let $a_1, a_2, \ldots, a_m$ be the attributes of $A$. The attribute sequence sorted by their significance in descending order is defined as \cite{Jiang2009}
\begin{equation}
    \begin{aligned}
    AQ=\langle a_1^\prime,a_2^\prime,\ldots,a_m^\prime\rangle,
    \end{aligned}
\end{equation}
where the attribute $a^{\prime}_i$ is better than the attribute $a_{i+1}^\prime$ with $Sig(\{a^{\prime}_i\}) \geq Sig(\{a_{i+1}^\prime\})$.

\noindent \textbf{Definition 12} (attribute subset sequence) Let $AQ=\langle a_1^\prime,a_2^\prime,\ldots,a_m^\prime\rangle$ be the attribute sequence sorted by their significance. The attribute subset sequence is defined as
\begin{equation}
    \begin{aligned}
    ASQ=\langle A_1,A_2,\ldots,A_m\rangle,
    \end{aligned}
\end{equation}
where $A_i$ is an attribute subset consisting of the top $i$ attributes in the attribute sequence $AS$, denoted as 
$A_i=\{a_j^\prime\ |\ a_j^\prime\in AS\ {\rm and}\ j\le i\}$.

\noindent \textbf{Definition 13} (outlier score) Let $FIS=(U,A,V,f)$ be a fuzzy information system. For any sample $x \in U$, the outlier score of $x$ is defined as
\begin{equation}
    \begin{aligned}
    S(x)=1-\frac{1}{|ASQ|}\sum_{A_i\in ASQ}{Sig(A_i)\frac{|\widetilde{[x]}_{A_i}|}{|U|}}.
    \end{aligned}\label{e19}
\end{equation}

In the definition, the outlier score of a sample is determined by the significance of attribute subsets and the fuzzy granule with density. When the size of the fuzzy granule with density is smaller, the sample is more likely to be an outlier. With the proposed measure for sample outlier score, a novel FRS-based outlier detection method is presented. The procedure for identifying outliers is described in Algorithm 1.

\begin{figure}[!ht]
    \label{FRDOD}
    \renewcommand{\algorithmicrequire}{\textbf{Input:}}
    \renewcommand{\algorithmicensure}{\textbf{Output:}}
  
    \begin{algorithm}[H]
        \caption{A fuzzy granule density-based outlier detection method}
        \LinesNumbered 
        \KwIn{A fuzzy information system $FIS = (U, A, V, f)$, similarity threshold $\delta$, and weighted parameter $\lambda$.}
        \KwOut{An outlier score vector $S$.}
        \For{\rm{each} $a \in A$}{
             Compute the fuzzy granule with density $\widetilde{[x]}_{{a}}$ for each sample and the significance $Sig({a})$;
        }
            Determine the attribute sequence $AQ$ and attribute subset sequence $ASQ$ according to the significance of each attribute;
            
            \For{\rm{each attribute subset} $A_i \in ASQ$}{
                Compute the significance of attribute subset $Sig(A_i)$;
        }
        \For{\rm{each}  $x \in U$}{
                Compute the outlier score $S(x)$;
        }
    \end{algorithm}
\end{figure}

Algorithm 1 first calculates the significance of each attribute, based on which attribute sequence and attribute subset sequence are determined. Then, the significance of each attribute subset within the sequence is computed. Finally, the outlier score of each sample is calculated by averaging the significance and fuzzy granule with density over all possible attribute subsets.

Assuming that a dataset has $|U|$ samples and $|A|$ attributes, the time complexity of computing fuzzy granule with density and attribute significance is {$O(|A||U|^2)$}. With the sorted attribute sequence and attribute subset sequence, the time complexity of computing the significance of all attribute subsets is at most {$O(|A||U|^2)$}. For each sample, it takes the time of $O(|A||U|)$ to calculate its outlier score. Therefore, the overall time complexity of Algorithm 1 is {$O(|A||U|^2)$}, and the space complexity is $O(|U|^2)$.

\subsection{Multi-scale granular ball computing}
FRS-based outlier detection is an effective distance-based method. Nevertheless, distance-based outlier detection methods encounter challenges when dealing with group outliers. Multi-view learning that leverages information from different views has been shown to be a useful technique for improving model performance. To effectively identify different types of outliers, we employ the technique of granular-ball computing to generate multi-scale views of data.

\noindent \textbf{Definition 14} (center and radius of granular ball) Let $GBS=\{GB_1, GB_2,\ldots, GB_n\}$ be a set of granular balls over all samples $U$. For any granular ball $GB_i \in GBS$, the center and radius of $GB_i$ are defined as \cite{Xia2022a}
\begin{equation}
    \begin{aligned}
        & c_i=\frac{1}{|GB_i|}\sum_{x \in GB_i} x,\\
        & r_i=\max_{x \in GB_i}\Vert x - c_i\Vert_2.
    \end{aligned}
\end{equation}

\noindent \textbf{Definition 15} (fuzzy similarity degree between granular balls) Let $GBS=\{GB_1, GB_2, \ldots, GB_n\}$ be a set of granular balls generated from fuzzy information system $FIS=(U, A, V,f)$. For any two granular balls {$GB_i, GB_j \in GBS$}, the fuzzy similarity degree with respect to an attribute $a \in A$ is defined as
\begin{equation}
\footnotesize
    \begin{aligned}
    R_{a}(GB_i,GB_j) =
        \begin{cases}
            1-dis_a(GB_i,GB_j), & dis_a(GB_i,GB_j)\le\epsilon_a,\\
            0,& {\rm otherwise, }
        \end{cases}
    \end{aligned}
\end{equation}
\noindent where $$dis_a(GB_i, GB_j)=\max(\vert f_a(c_i)-f_a(c_j) \vert - |r_i^{\frac{1}{|A|}} + r_j^{\frac{1}{|A|}}|,0),$$ $\epsilon_a$ is the adjustable radius, which is calculated by $\epsilon_a=\frac{std(a)}{\delta}$, and $std(a)$ is the standard deviation of attribute values on $a$ and $\delta$ is an adjustable parameter. 

\begin{figure*}[!ht]
	\footnotesize
	\centering
            \subfigure[]{\includegraphics[scale=0.051,trim=280 180 320 330, clip]{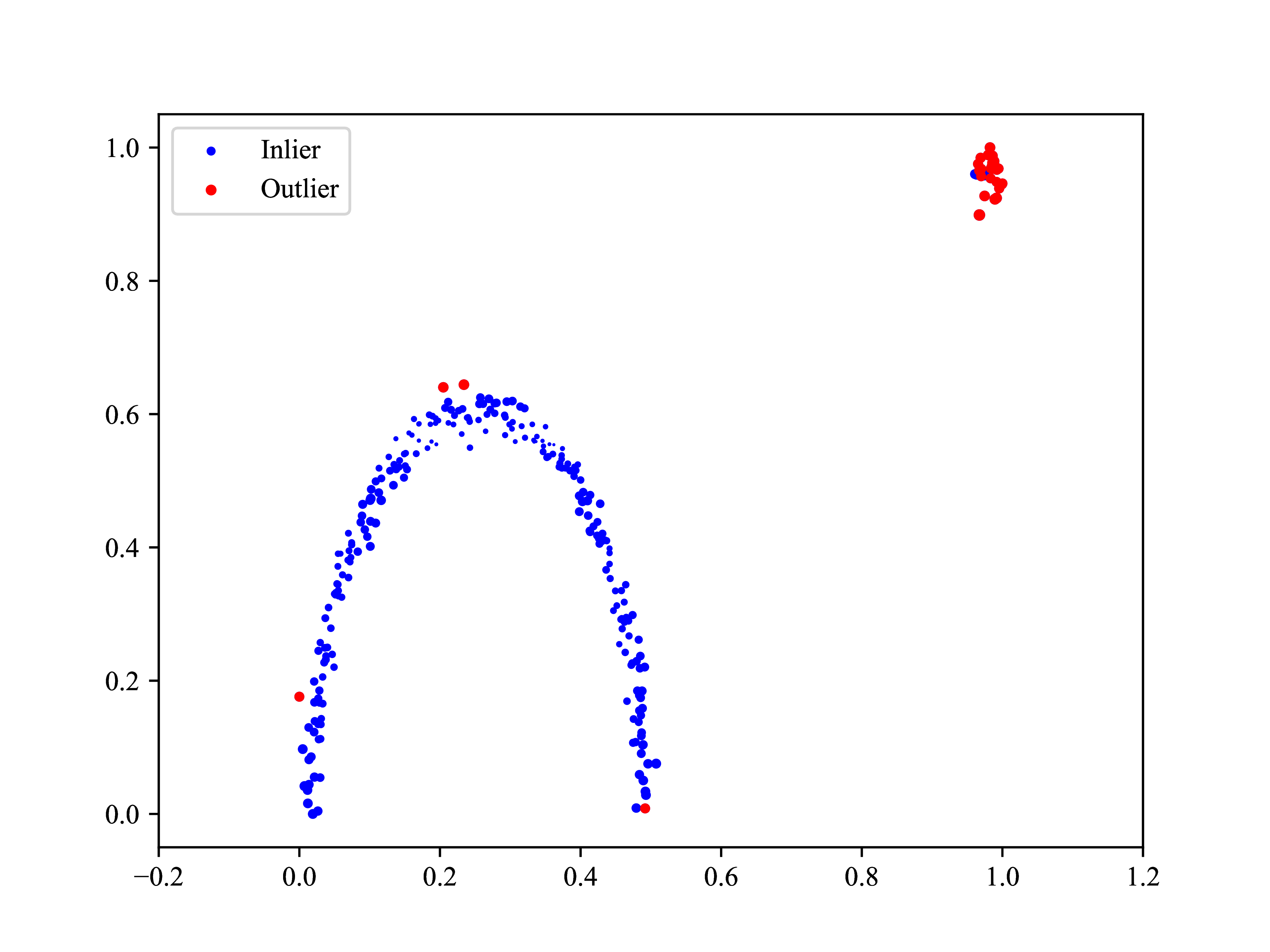}}
		\subfigure[]{\includegraphics[scale=0.051,trim=280 180 320 330, clip]{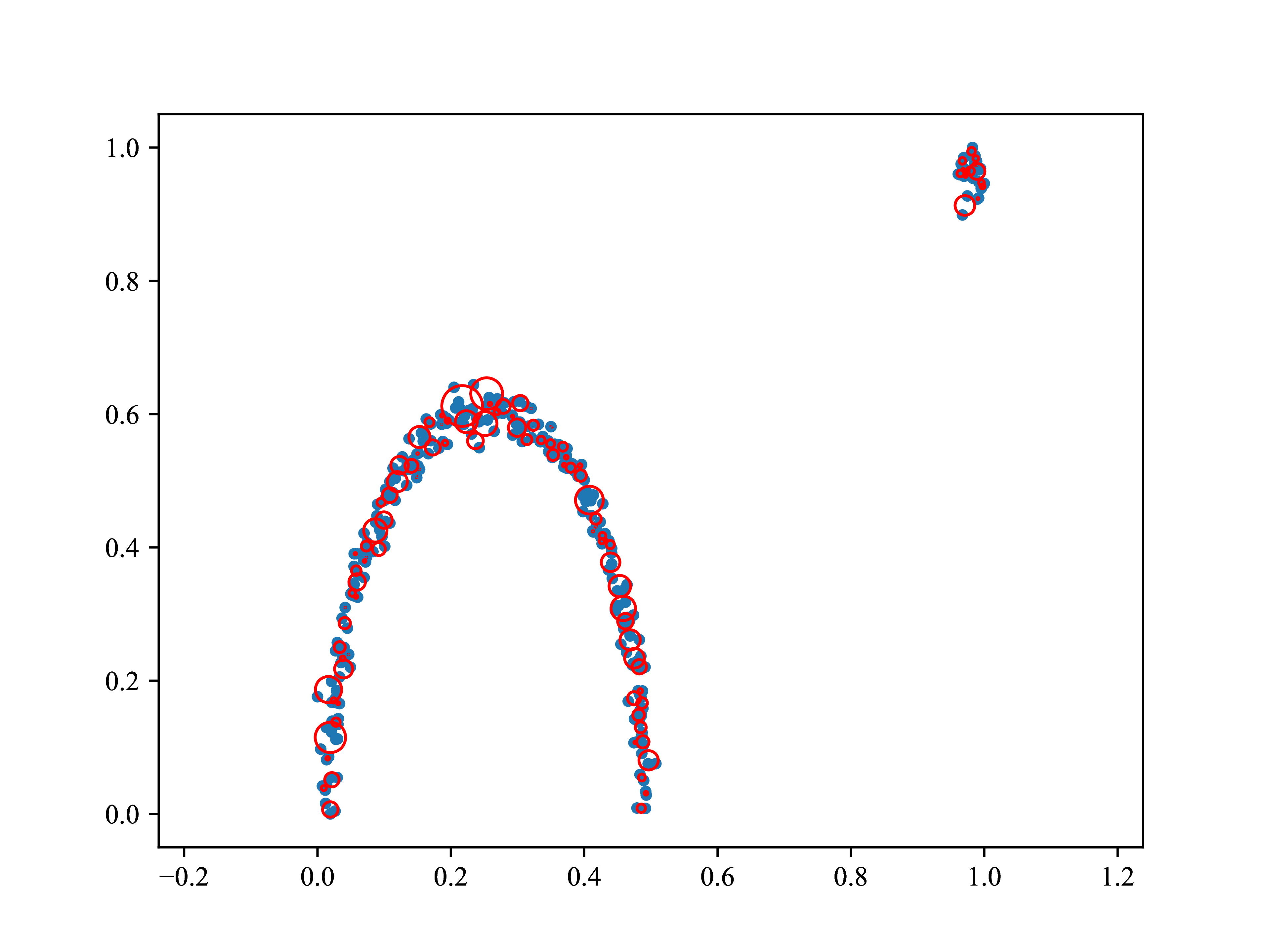}}
		\subfigure[]{\includegraphics[scale=0.051,trim=280 180 320 330, clip]{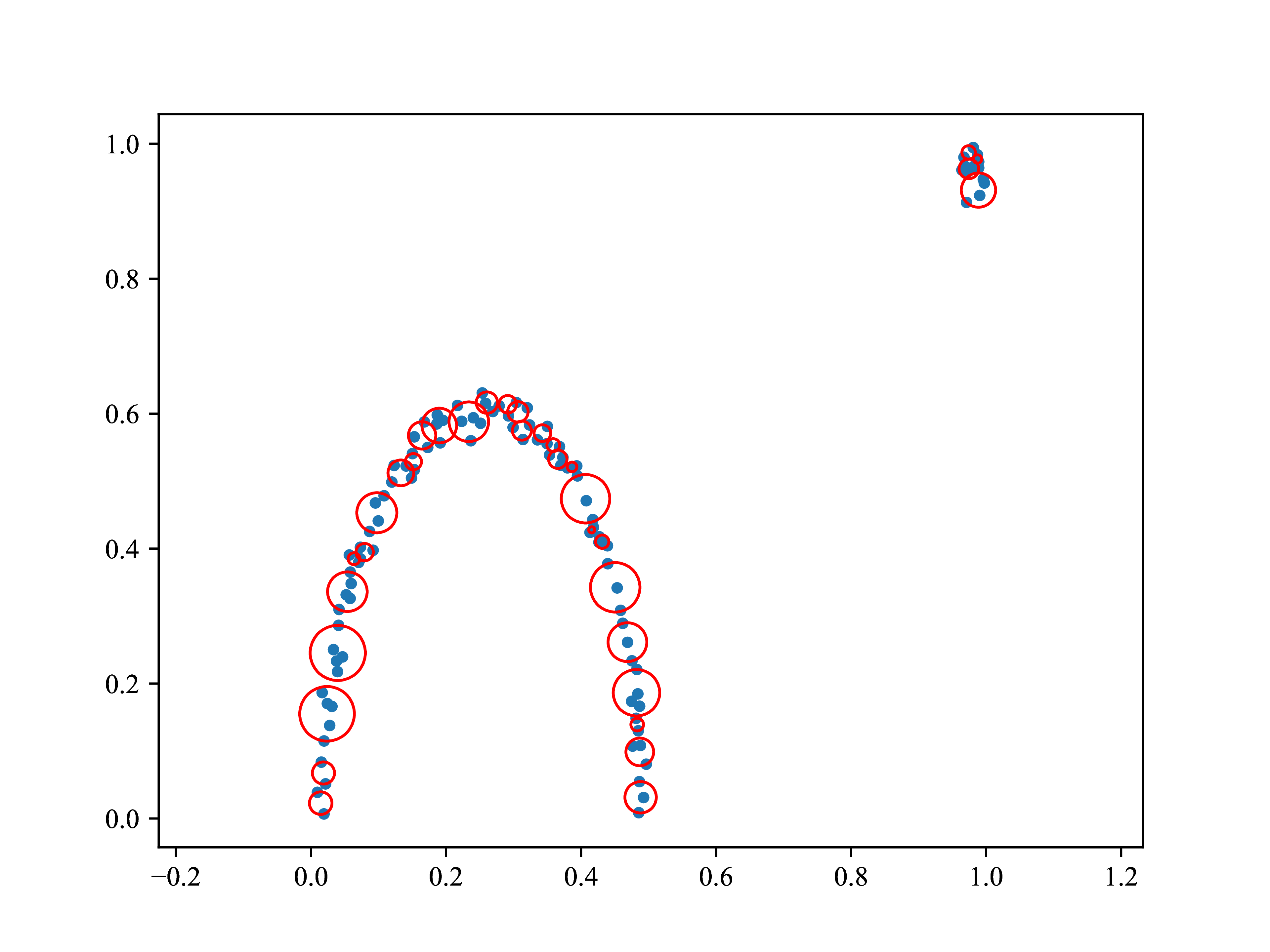}}
  		\subfigure[]{\includegraphics[scale=0.051,trim=280 180 320 330, clip]{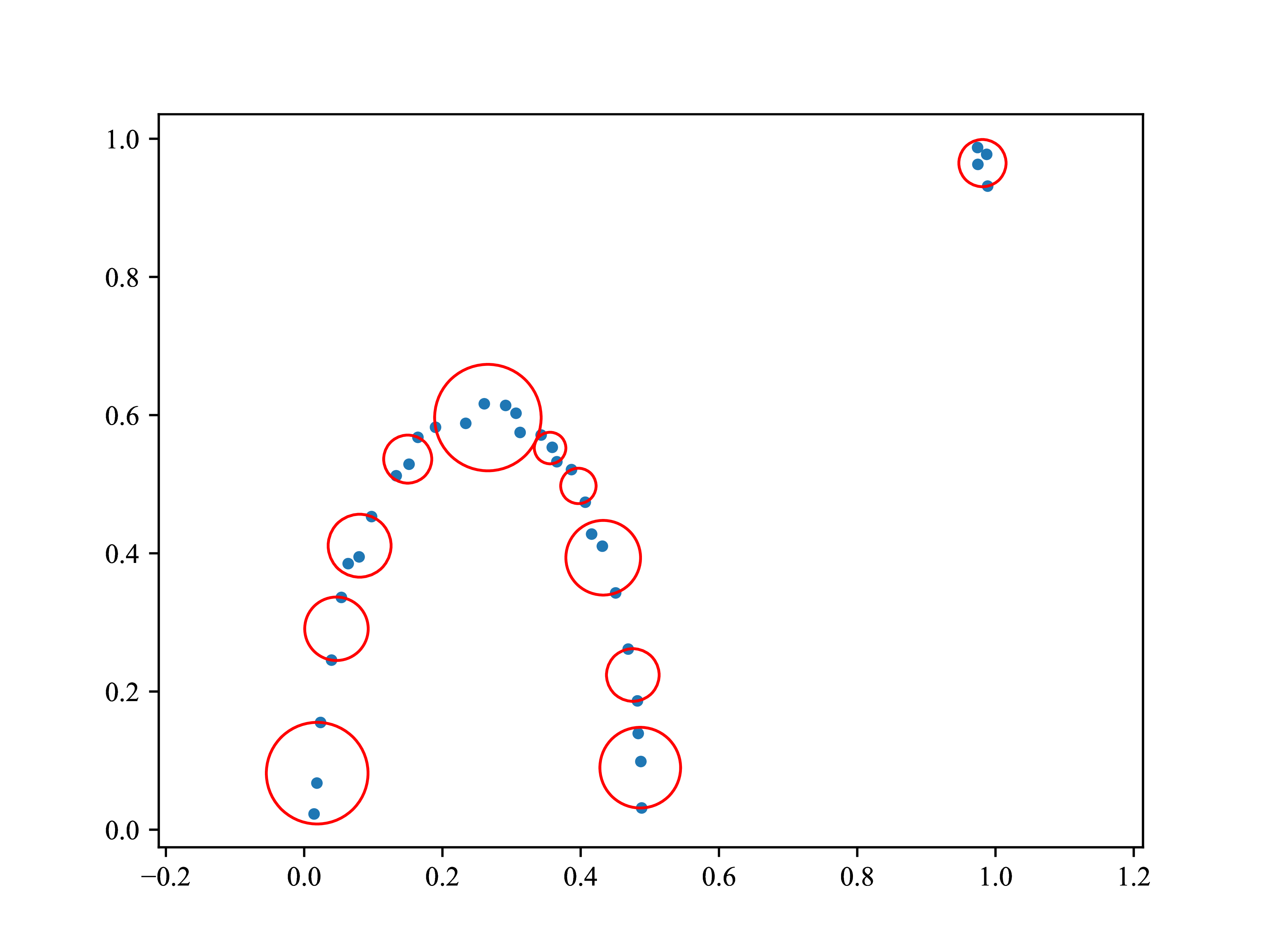}}
		\subfigure[]{\includegraphics[scale=0.051,trim=280 180 320 330, clip]{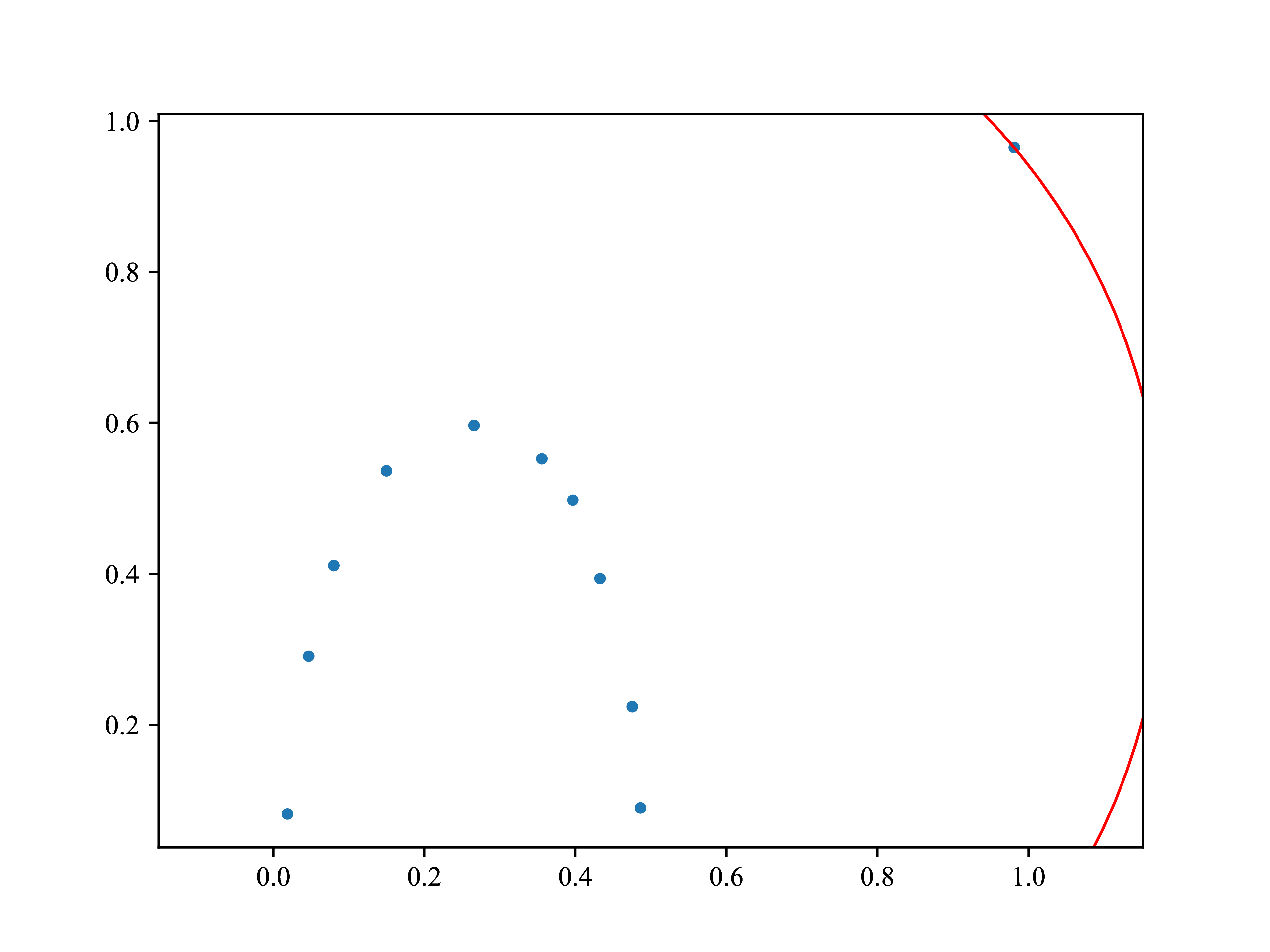}}
		\subfigure[]{\includegraphics[scale=0.051,trim=280 180 320 330, clip]{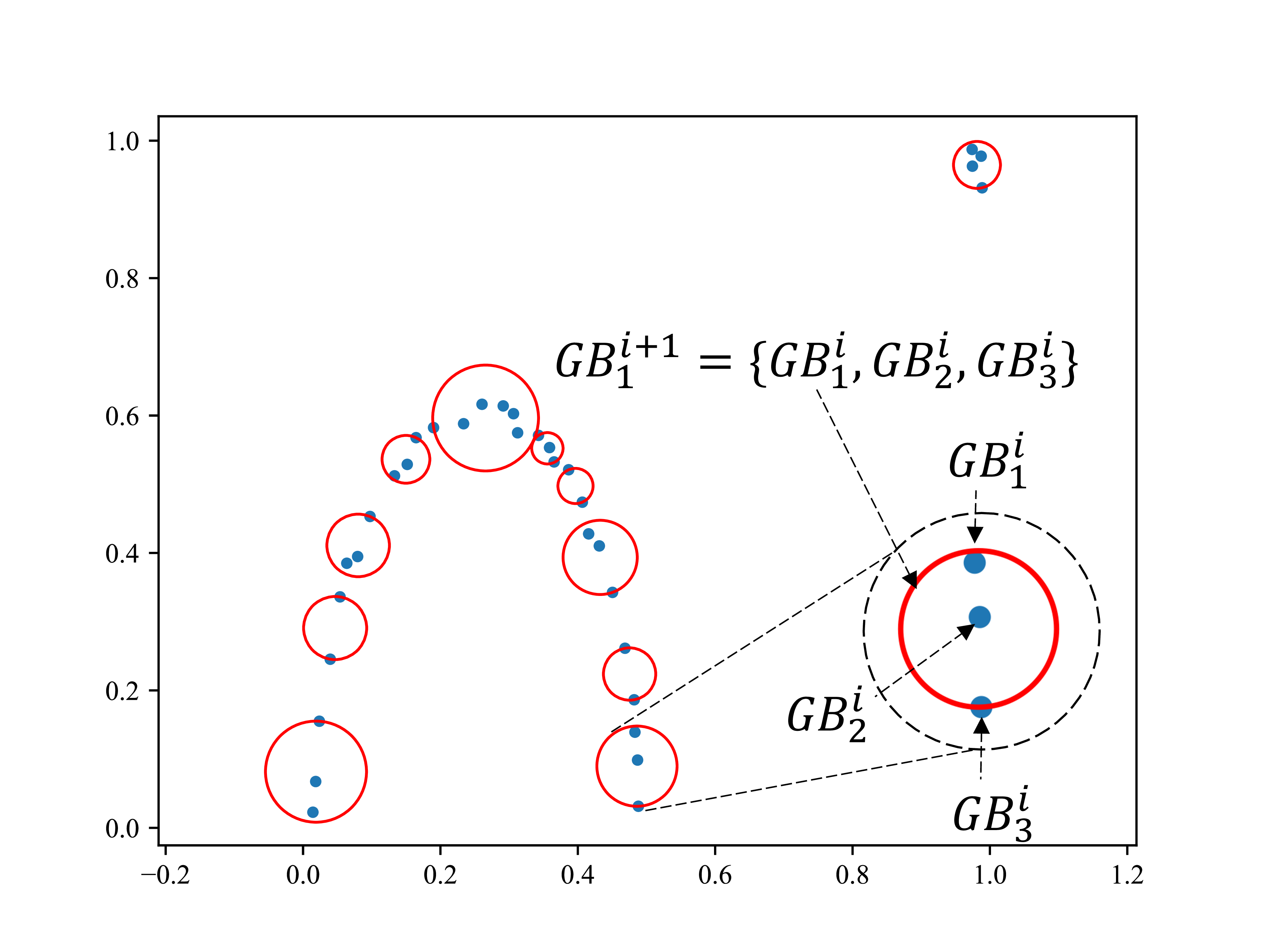}}
		\par
	\caption{Multi-scale view generation with granular ball computing on a synthetic dataset. a) the original view and outlier detection results; b) the 2nd scale view; c) the 3rd scale view; d) the 4th scale view; e) the last scale view; {(f) granular ball updating. In the process of multi-scale view generation, the granular balls $GB_1^i$, $GB_2^i$, and $GB_3^i$ in the $i$-th scale view are merged into a coarser granular ball $GB_1^{i+1}$ for the $(i+1)$-th scale view.}}\label{f5}
\end{figure*}

{Based on the fuzzy similarity degree of granular balls and the idea of granular ball computing, different scales of data can be generated successively (see Fig. \ref{f5})}. Specifically, the original data is considered as the finest scale view where each granular ball contains only one sample, while the higher scale views can be derived from preceding generated lower scale views by enlarging granular balls to enclose similar samples. The process of view generation is terminated at the coarsest scale view where all samples are grouped in a single granular ball. The procedure of generating multi-scale views can be described as Algorithm 2.

\begin{figure}[!ht]
    \label{MVGGC}
    \renewcommand{\algorithmicrequire}{\textbf{Input:}}
    \renewcommand{\algorithmicensure}{\textbf{Output:}}
    \setcounter{AlgoLine}{0}
    \begin{algorithm}[H]
        \caption{Multi-scale view generation based on granular-ball computing}
        \LinesNumbered 
        \KwIn{A fuzzy information system $FIS=(U, A, V, f)$.}
        \KwOut{A family of granular ball sets $GBSV$.}
            Treat the original data $U$ as a basic set of granular balls $GBS_1$, with each granular ball containing only one sample;
            
            Initialize $GBSV\gets\{GBS_1\}$ and $k \gets 1$;
            
            \While{$|GBS_k|\neq 1$}{
                Compute the fuzzy similarity degree between granular balls generated in the prior scale view $GBS_k$;
                
                Generate a coarser set of granular balls $GBS_{k+1}$ based on the idea of granular ball computing;
                
                $GBSV \gets GBSV \cup \{GBS_{k+1}\}$ and $k\gets k + 1$;
            }
    \end{algorithm}
\end{figure}

Algorithm 2 begins with the original finest scale view where each sample is considered as a granular ball. Then, it iteratively generates a coarser scale view by combining granular balls within the prior finer scale view until all samples are grouped into one granular ball, at which the coarsest scale view is obtained. According to the analysis of the work \cite{Xia2022a}, the time complexity of generating single-scale granular balls is less than $O(|U|\log|U|)$. {It is found that the number of samples $|U|$ and multi-scale views $|GBSV|$ satisfies $|GBSV| \sim \log |U|$}\footnote{Theoretical analysis and experimental validation are provided on the GitHub repository (\url{https://github.com/Xiaofeng-Tan/MGBOD}).}. {Thus, the overall time complexity of multi-scale view generation is $O(|U|(\log|U|)^2)$, and the space complexity is $O(|U|^2)$.}

\subsection{Multi-scale granular balls-based outlier detection}
The multi-scale views of the data inherently help outlier detection methods to identify different types of outliers. The original view has the finest granular balls that contain only one sample, and more detailed information about the data is provided, which facilitates the detection of local and global outliers. Conversely, in the coarser scale views, more samples are aggregated into granule balls, and group outliers that are difficult to identify in the original view become detectable. To effectively detect different types of outliers, we introduce three-way decision and weighted SVM to fuse outlier detection results from multi-scale views. 

By using the fuzzy similarity degree between granular balls, multi-scale views of the data can be generated, and the proposed FRS-based outlier detection method can be applied to different scale views directly, and the outlier score of each sample is determined by that of the granular ball to which it belongs.

\noindent \textbf{Definition 16} (outlier probability in each view)  Let $t$ be the proportion of outliers. For any sample $x \in U$, the outlier probability $P_{k}(x)$ in the $k$-th scale view is defined as
\begin{equation}
    \begin{aligned}
    P_{k}(x) =
        \begin{cases}
            \frac{S_{k}(x)-S_{k}^o}{2(\max(S_k)-S_k^o)}+\frac{1}{2}, & S_k(x)\geq S_k^o,\\
            \frac{S_k(x)-\min(S_k)}{2(S_k^{o+1}-\min(S_k))}, & {\rm otherwise,}
        \end{cases}
    \end{aligned}\label{e26}
\end{equation}
where the $S_k(x)$ represents the outlier score of sample $x$ in the $k$-th scale view, $S_k^o$ denotes the $o$-th largest value within the outlier score vector $S_k$, and $o$ is the number of estimated outliers according to the proportion of outliers, which is computed by ceiling up the product of $t$ and $|U|$, denoted as $\lceil t|U| \rceil$.

After probability mapping, the outlier scores of samples are normalized into the range of $[0, 1]$, where the top $o$ outlier scores are mapped into the range of $[0.5, 1]$. By transforming outlier scores into outlier probability, the quantification of samples being outliers becomes more semantic, and outlier detection results from different scale views can be aligned for fusion.

\noindent \textbf{Definition 17} (weight of each view)  Let $P_{k}$ be the outlier probability vector in the $k$-th scale view. The weight of the $k$-th scale view is defined as
\begin{equation}
    {\nu}_k=1-\frac{\sum_{x \in U}{H_k(x)}}{|U|},
\end{equation}
where
\begin{equation}
\small
    H_k(x)=-P_k(x) \log P_k(x)-(1-P_k(x)) \log (1-P_k(x)),
\end{equation}
and $P_k(x)$ denotes the outlier probability of the sample $x$ in the $k$-th scale view.

\noindent \textbf{Definition 18} (weight of each sample)  Let $P_{k}$ and ${\nu}_k$ be the outlier probability vector and weight of the $k$-th scale view, respectively. For any sample $ x \in U$, the weight of the sample $x$ is defined as 
\begin{equation}
    \begin{aligned}
            \mu(x)=1-\frac{\sum_{k=1}^{K}{\nu}_k\cdot H_k(x)}{K},
    \end{aligned}
\end{equation}
where $K$ is the number of generated multi-scale views.

The weight of the $k$-th view is defined by averaging information entropy over the outlier probability distribution of each sample in the corresponding view. When outliers are indistinguishable from normal samples, the outlier probability of samples tends to be near 0.5. In this case, the average information entropy of samples is approximated to 1, and the weight is close to 0, indicating that the view has a very limited contribution to outlier detection. While the weight of each sample reflects the uncertainty in identifying the sample as an outlier. When the outlier probability of a sample is near 0.5, the lower the certainty of detecting the sample as an outlier or inlier and the smaller weight of the sample.

\noindent \textbf{Definition 19} (weighted outlier probability) Let $P_{k}$ and ${\nu}_k$ be the outlier probability vector and weight of the $k$-th scale view, respectively. For any sample $ x \in U$, the weighted outlier probability ${P}(x)$ over all scale views is defined as
\begin{equation}
    \begin{aligned}
    {P}(x)=\frac{
    \sum_{k=1}^{K}{{\nu}_k \cdot P_k(x)}}{\sum_{k=1}^{K}{\nu}_k},
    \end{aligned}\label{e28}
\end{equation}
where $P_k(x)$ denotes the outlier probability of the sample $x$ in the $k$-th scale view. 

\noindent \textbf{Definition 20} (confidence thresholds) Let $t$ be the proportion of outliers and $OP(i)$ be the $i$-th largest probability of being an outlier in ascending order. The upper and lower confidence thresholds $\alpha$ and $\beta$ are defined as
\begin{equation}
    \begin{aligned}
            \alpha&=OP(\lceil|U|(1-t+\Delta t))\rceil),\\
            \beta&=OP(\lceil|U|(1-t-\Delta(1-t))\rceil),
    \end{aligned}
\end{equation}
where $\Delta$ is a parameter that determines the size of the asymmetrical neighborhood, and $\lceil.\rceil$ denotes the ceil function that rounds up a number to the smallest integer. In this study, $\Delta$ is set to 0.7. 

\noindent \textbf{Definition 21} (three-way regions) Let ${P}$ be the weighted outlier probability vector over all scale views and $\alpha$, $\beta$ be the confidence thresholds. The positive, boundary, and negative regions of all samples are defined as 
\begin{equation}
    \begin{aligned}
            {POS}_{\beta}^{\alpha}(U)&= \{x \in U|{P}(x)\geq\alpha\}, \\
            {BND}_{\beta}^{\alpha}(U)&= \{ x \in U|\beta<{P}(x)<\alpha\}, \\
            {NEG}_{\beta}^{\alpha}(U)&= \{ x\in U|{P}(x) \le\beta\},
    \end{aligned}
\end{equation}
where ${P}(x)$ denotes the weighted outlier probability of the sample $x$. 

Based on the idea of tri-partition in the three-way decision \cite{Zhang2017}, all samples can be categorized into three regions in terms of their weighted outlier probability. Samples with the weighted outlier probability greater than $\alpha$ can be confidently identified as outliers. Conversely, samples with the outlier probability lower than $\beta$ can be certainly considered inliers. While samples with outlier scores falling between $\beta$ and $\alpha$ are regarded as uncertain since a confident decision cannot be made. To further identify outliers from the uncertain region accurately, we propose a weighted SVM using reliable outliers in the positive region and inliers in the negative region, and the objective function can be formulated as

\begin{equation}
    \begin{aligned}
            & \min_{\boldsymbol{w},b,\boldsymbol{\xi}}\ \frac{1}{2}\Vert \boldsymbol{w}\Vert^2_2+C^{+}\sum_{x_i\in {POS}_{\beta}^{\alpha}(U)} \mu(x_i) \xi_i \\
            &+C^{-}\sum_{x_j\in {NEG}_{\beta}^{\alpha}(U)} \mu(x_j) \xi_j ,\\ 
            & {\rm s.t.}\; y_k(\boldsymbol{w}x_k+b)\geq 1- \xi_k,\; \xi_k\geq 0, k=1, 2, \cdots, l,
    \end{aligned}\label{e34}
\end{equation}
where $\mu(x)$ denotes the weight of the sample $x$, $\xi$ signifies slack variable, $C^+$ and $C^-$ stands for the trade-off parameters for outliers and inliers, respectively, $y_k$ denotes the label of the sample $x_k$, with the value of $+1$ for outliers and $-1$ for inliers, and $l$ is the number of reliable outliers and inliers, denoted as $l=|{POS}_{\beta}^{\alpha}(U) \cup {NEG}_{\beta}^{\alpha}(U)|$.

In the objective function, to address the problem of imbalance between outliers and inliers, the constrain $\frac{C^+}{C^-}=\frac{t}{1-t}$ is imposed on the parameters $C^+$ and $C^-$, where $t$ is the proportion of outliers. By using the method of Lagrange multiplier, the objective function (\ref{e34}) can be converted into the following dual problem 
\begin{equation}
    \begin{aligned}
    & \min_{\boldsymbol{\eta}}\frac{1}{2}\sum_{x_i\in {POS}_{\beta}^{\alpha}(U)}\sum_{x_j\in {NEG}_{\beta}^{\alpha}(U)}{\eta_i\eta_jy_iy_jx_i^{\rm{T}}x_j}-\sum_{k=1}^{l}\eta_k,\\
    & {\rm s.t.}\sum_{k=1}^{l}\eta_ky_k=0,\\
    & 0\le\eta_i\le \mu(x_i){C}^+, 0\le\eta_j\le \mu(x_j){C}^-.
    \end{aligned}\label{e35}
\end{equation}
The optimization of (\ref{e35}) is a quadratic programming problem and can be solved using the algorithm of Sequential Minimal Optimization (SMO) \cite{Platt1998}.

By using the trained SVM classifier, samples within the uncertain region can be further determined the probability of being outliers, and a refined outlier probability vector is finally generated. The process of multi-scale outlier detection with granular balls can be described by Algorithm 3.

\begin{figure}[!ht]
    \label{FRDODMV}
    \renewcommand{\algorithmicrequire}{\textbf{Input:}}
    \renewcommand{\algorithmicensure}{\textbf{Output:}}
    \setcounter{AlgoLine}{0}
    \begin{algorithm}[H]
        \caption{Fuzzy granule density-based outlier detection with multi-scale granular balls.}
        \LinesNumbered 
        \KwIn{A fuzzy information system $FIS=(U, A, V, f)$, similarity threshold $\delta$, weighted parameter $\lambda$, and the proportion of outliers $t$.}
        \KwOut{An outlier probability vector$\widetilde{P}$.}
        
             Perform multi-scale view generation on $U$ using \textbf{Algorithm 2} and obtain a set of granular ball views $GBSV$;
             
            \For{\rm{each granular ball view} $GBS_k \in GBSV$}{
                Compute the outlier score vector $S_k$ using \textbf{Algorithm 1};
                
                Map the outlier score vector $S_k$ into the outlier probability vector $P_k$;
                
                Compute the weight $\nu_k$ of the $k$-th scale view;
                
            }
            
        Compute the weighted outlier probability vector $P$ and the weight $\mu (x)$ of each sample;
        
        Determine the three-way regions using the calculated confidence thresholds $\alpha$ and $\beta$;
        
        Train the weighted SVM using reliable outliers in the positive region and inliers in the negative region;
        
        Compute the final outlier probability vector $\widetilde{P}$ by the trained SVM;
        
    \end{algorithm}
\end{figure}

Algorithm 3 generates a set of data views using the multi-scale granular ball generation method described in Algorithm 2. For each granular ball view,  the outlier score vector of samples is computed and mapped into the outlier probability vector, based on which the weight of the view is calculated. Then, the weighted outlier probability vector as well as the sample weights are computed by fusing the results from different views. Subsequently, a weighted SVM is trained on the reliable outliers and inliers determined by the theory of three-way decision with the calculated confidence thresholds. Finally, a refined outlier probability vector is obtained by using Platt scaling to transform the outputs of the weighted SVM into a probability distribution over outlier and inlier classes.

The algorithm involves three stages: the generation of multi-scale views, the computation of sample outlier scores, and the training of weighted SVM. Assuming that a dataset has $|U|$ samples and $|A|$ attributes. As discussed above, the time complexity of generating single-scale granular balls is $O(|U|\log|U|)$, while the time complexity of the proposed outlier detection method {for each view is $O(|A||U|^2)$. The relationship between the number of samples $|U|$ and multi-scale views $|GBSV|$ satisfies $|GBSV| \sim \log |U|$, and thus the overall time complexity of outlier detection for all views is $O(|A||U|^2 \log |U|)$.} For the training of weighted SVM, the time complexity is $O(|A||U|^2 )$ when using the SMO algorithm. Thus, the overall time complexity of Algorithm 3 is {$O(|A||U|^2\log |U|)$}, and the space complexity is $O(|U|^2)$.

\section{Experiments}\label{section 4}
In this section, we first examine the effectiveness of the proposed method in identifying different types of outliers using three synthetic datasets. Then, we conduct extensive experiments to compare the proposed method with other state-of-the-art methods. Finally, we perform the ablation experiments on the proposed multi-scale outlier detection method. All methods were implemented in the development platform of Visual Studio Code with Python 3.9, and all experiments were carried out on a computer running the Ubuntu 20.04.1 operating system with an Intel(R) Core (TM) i7-10700 CPU @ 2.90GHz, and 32 GB RAM.

\subsection{Datasets and evaluation metrics}

Twenty datasets obtained from publicly accessible repositories were selected for experiments, which are often used in outlier detection tasks. In some of datasets, outliers are formed by randomly downsampling specific classes, while the other classes are preserved as normal samples. The information about these datasets are summarized in Table \ref{table2}\footnote{Refer to the GitHub repository (\url{https://github.com/Xiaofeng-Tan/MGBOD}) for more detailed information}, where the second column reports the name and abbreviation of the selected datasets, with the corresponding data source, the third to fifth columns show the number of attributes, samples, and outliers, respectively, and the last column indicates the proportion of outliers.

\begin{table}[!ht]
	\footnotesize
	\caption{The investigated datasets. 
        \label{datasets}}
	\begin{center}
		\setlength{\tabcolsep}{1.5mm}
		\begin{tabular}{clcccc}
			\toprule
			No. & Datasets (Abbr.)$^{\rm{source}}$ & $|A|$ & $|U|$ & $|O|$ & $t$\\ 
			\midrule
        1 & Arrhythmia (Arrhyth)$^{\rm{1}}$ & 274 & 452 & 66 & 0.1460 \\ 
        2 & Autos (Autos)$^{\rm{1}}$ & 25 & 205 & 25 & 0.1219 \\ 
        3 & Breastw (Breast)$^{\rm{2}}$ & 9 & 683 & 239 & 0.3499 \\ 
        4 & Cardio (Cardio)$^{\rm{1}}$ & 21 & 1831 & 176 & 0.0961 \\ 
        5 & Cardiotocography (Cardioto)$^{\rm{2}}$ & 21 & 2114 & 466 & 0.2204 \\ 
        6 & Chess (Chess)$^{\rm{1}}$ & 36 & 1896 & 227 & 0.1197 \\ 
        7 & Hepatitis (Hepat)$^{\rm{2}}$ & 19 & 80 & 13 & 0.1625 \\ 
        8 & Ionosphere (Iono)$^{\rm{1}}$ & 34 & 249 & 24 & 0.0963 \\ 
        9 & Iris (Iris)$^{\rm{1}}$ & 4 & 111 & 11 & 0.0991 \\ 
        10 & Mammography (Mammo)$^{{\rm{2}}}$ & 6 & 11183 & 260 & 0.0232 \\ 
        11 & MVTec-AD\_carpet (Carpet)$^{\rm{2}}$ & 512 & 397 & 89 & 0.2241 \\ 
        12 & MVTec-AD\_metal\_nut (Metal)$^{\rm{2}}$ & 512 & 335 & 93 & 0.2776 \\ 
        13 & MVTec-AD\_pill (Pill)$^{\rm{2}}$ & 512 & 434 & 141 & 0.3248 \\ 
        14 & Pendigits (Pen)$^{\rm{2}}$ & 16 & 6870 & 156 & 0.0227 \\ 
        15 & Satimage-2 (Sat)$^{\rm{2}}$ & 36 & 5803 & 71 & 0.0122 \\ 
        16 & SpamBase (Spam)$^{\rm{2}}$ & 57 & 4207 & 1679 & 0.3990 \\ 
        17 & Thyroid (Thyroid)$^{\rm{1}}$ & 28 & 9172 & 74 & 0.0081 \\ 
        18 & WDBC (WDBC)$^{\rm{1}}$ & 31 & 396 & 39 & 0.0985 \\ 
        19 & Wine (Wine)$^{\rm{2}}$ & 13 & 129 & 10 & 0.0775 \\ 
        20 & WPBC (WPBC)$^{\rm{2}}$ & 33 & 198 & 47 & 0.2374 \\ 
			\bottomrule
	\end{tabular}\label{table2}
	\end{center}
   \footnotesize{$^{\rm{1}}$ \url{https://github.com/BElloney/Outlier-detection}}\\
 \footnotesize{$^{\rm{2}}$ \url{https://github.com/Minqi824/ADBench}}
\end{table}
\color{black}

To evaluate the performance of the selected methods, the precision and recall rates were used as the evaluation metrics. Given the proportion of outliers $t$, an outlier probability threshold can be calculated by $\theta_t = OP(\lceil t|U|\rceil)$, where  $\lceil.\rceil$ denotes the ceil function, and $OP(i)$ stands for the $i$-th largest value in the final outlier probability vector $\widetilde{P}$. The precision and recall rates are defined as \cite{Campos2016}

\begin{equation}
    \begin{aligned}
        {\text{Precision}}(t)=\frac{|OS(t)\cap OS^o|}{|OS(t)|},
    \end{aligned}
\end{equation}

\begin{equation}
    \begin{aligned}
        {\text{Recall}}(t)=\frac{|OS(t)\cap OS^o|}{|OS^o|},
    \end{aligned}
\end{equation}
where $OS(t)$ is the set of identified outliers, denoted as $OS(t) = \{x | \widetilde{P}(x) > \theta_t\}$, and {
$OS^o$} is the set of ground-truth outliers. 

Moreover, the Receiver Operating Characteristic (ROC) curve and the Area Under the ROC Curve (AUROC) were employed as the evaluation metrics to comprehensively assess the performance of outlier detection, where the true positive rate $TPR(t)$ and false positive rate $TPR(t)$ are defined as \cite{Campos2016}
\begin{equation}
    \begin{aligned}
            TPR(t)=\frac{|OS(t)\cap OS^o|}{|OS^o|} ,
    \end{aligned}
\end{equation}

\begin{equation}
    \begin{aligned}
            FPR(t)=\frac{|OS(t)-OS^o|}{|U-OS^o|} ,
    \end{aligned}
\end{equation}

When varying the parameters $t$, a number of point pairs of $TPR(t)$ and $FPR(t)$ can be obtained and used to plot the ROC curve. Obviously, the outlier detection method achieves better performance when $TPR(t)$ is larger and $FPR(t)$ is smaller. To accurately quantify the performance of the outlier detection methods, the Area Under the ROC curve (AUROC) was employed. Given the final outlier probability vector $\widetilde{P}$, the AUROC can be calculated as \cite{Campos2016}
{
\begin{equation}
    \begin{aligned}
        {AUROC} = 1 - \frac{1}{|OS^o||U-OS^o|}\sum_{x_i\in OS^o}\sum_{x_j\in(U-OS^o)}\\
        \Big( \mathbb{I}\big(\widetilde{P}(x_i)>\widetilde{P}(x_j)\big)+0.5\mathbb{I}\big(\widetilde{P}(x_i)=\widetilde{P}(x_j)\big) \Big),
    \end{aligned}
\end{equation}}
where $\mathbb{I}(\cdot)$ denotes the indicator function.

\subsection{Performance on detecting various types of outliers}
\begin{figure*}[!ht]
	\begin{centering}
		\subfigure[\label{f6-a}]{\includegraphics[scale=0.058] {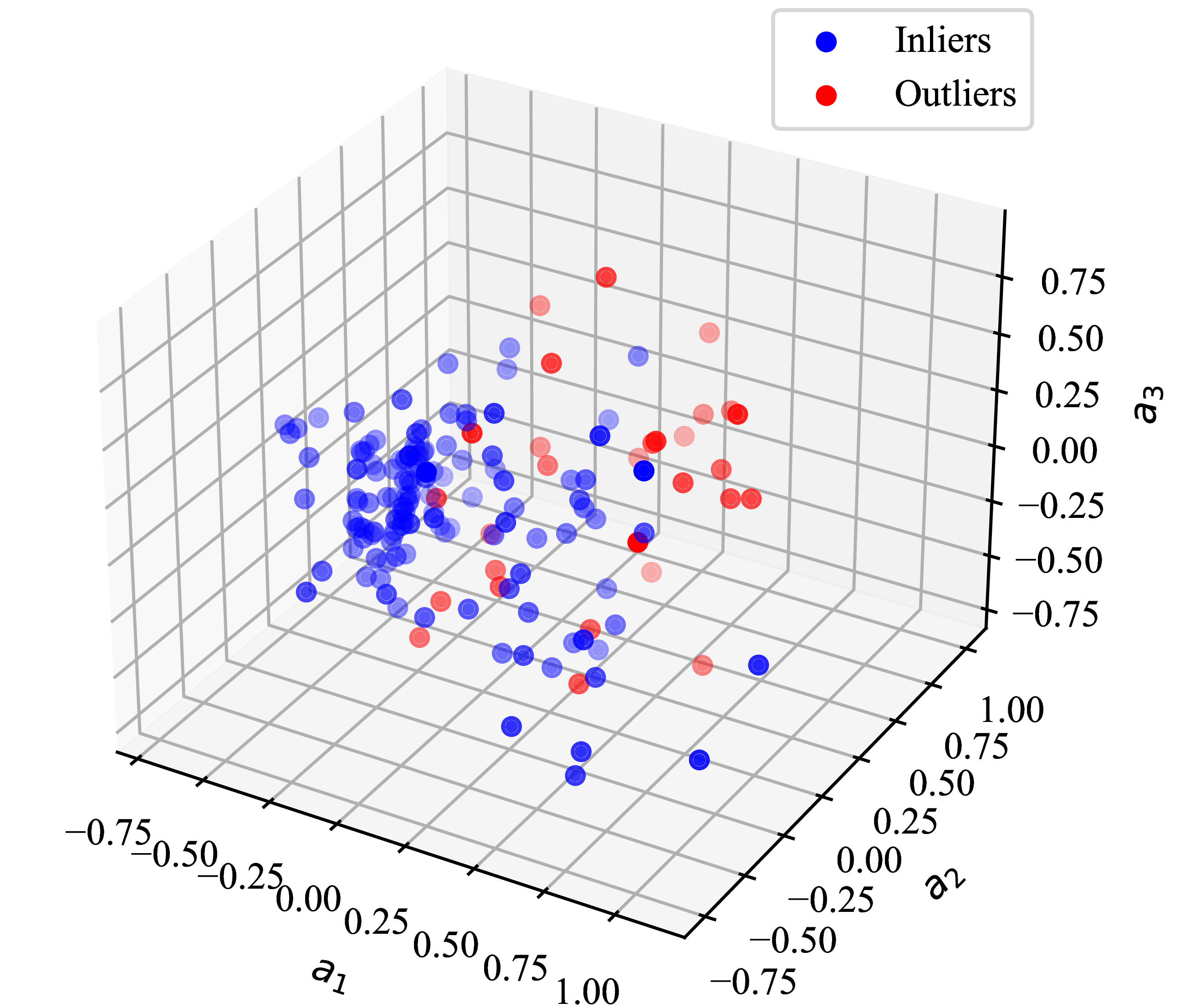}}
		\subfigure[\label{f6-b}] {\includegraphics[scale=0.058] {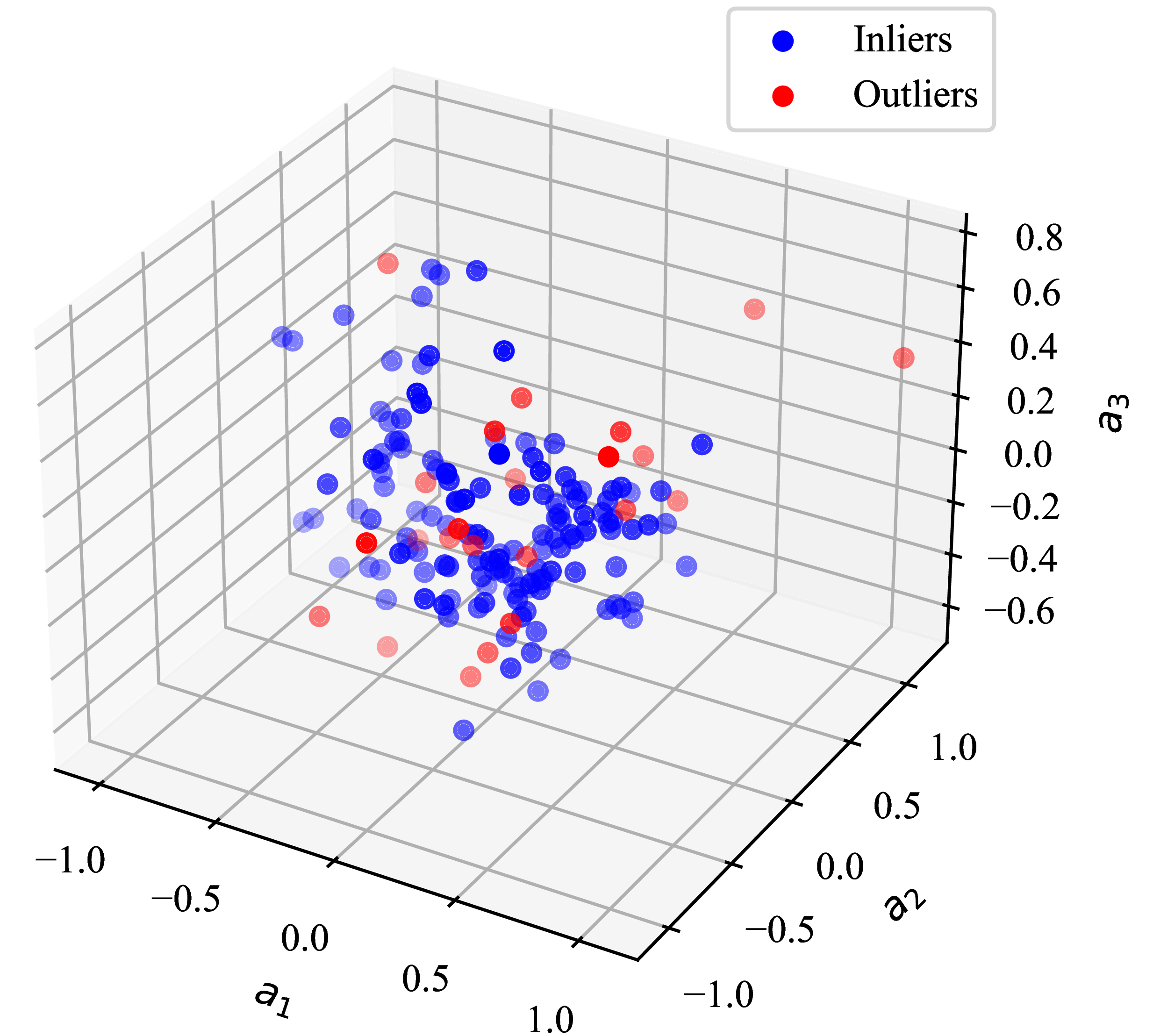}}
  		\subfigure[\label{f6-c}] {\includegraphics[scale=0.058] {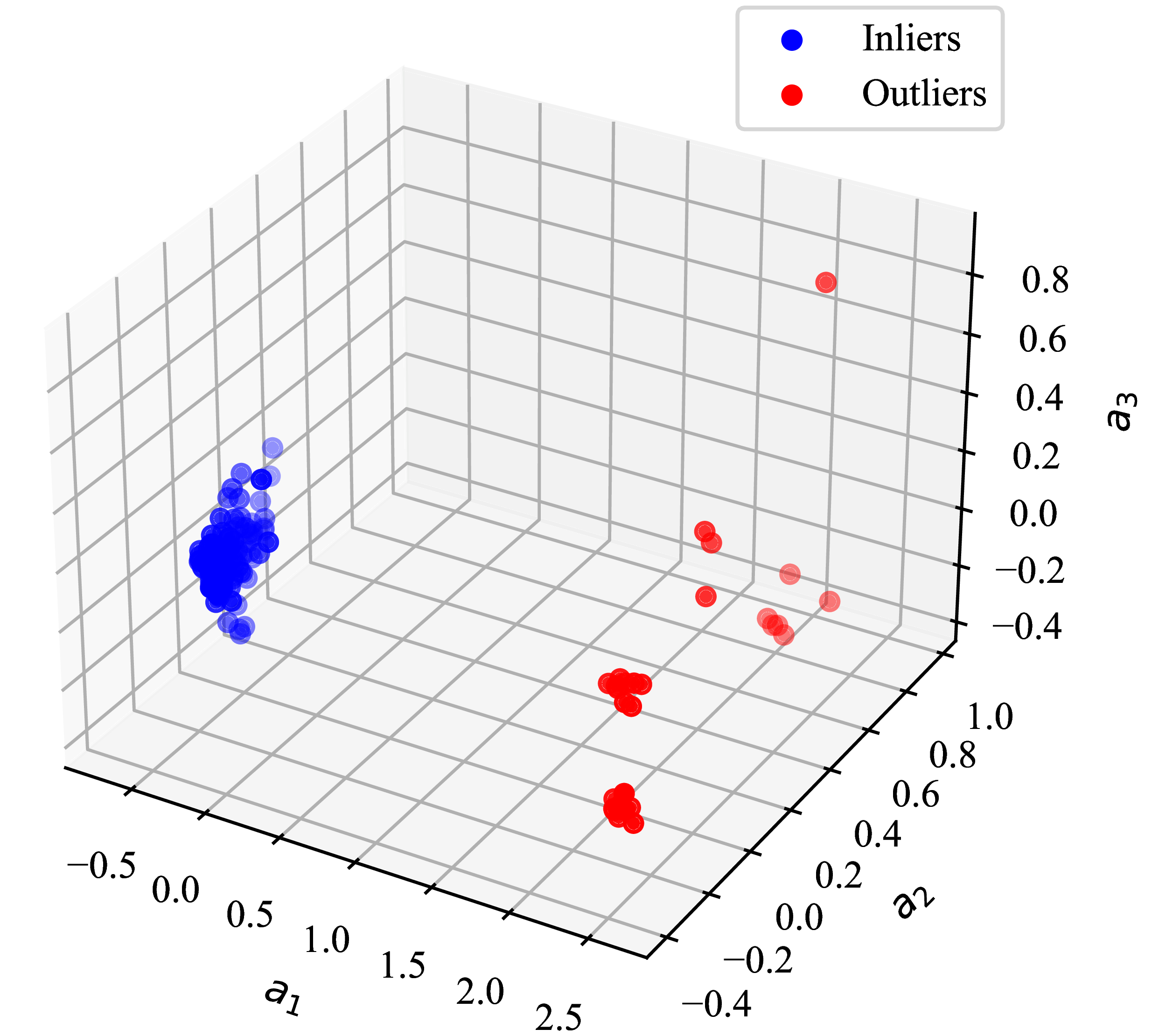}}
		\par\end{centering}
	\caption{The synthesized datasets visualized in 3-dimensional PCA space. a) Local outliers; b) Global outliers; c) Group outliers.}\label{f6}
\end{figure*}

To examine the effectiveness of the proposed FRS-based method in detecting different types of outliers, we synthesized three datasets by injecting local, global, and group outliers, respectively, which are generated from the dataset ``Hepat” using the method presented in  \cite{Han2022}. Fig. \ref{f6} visualizes the synthesized datasets and outlier detection results in 3-dimensional PCA space.

Table \ref{table3} shows the performance of both the FRS-based method and our proposed method on detecting local, global, and group outliers, where the precision rate and recall rate under different percentages of outliers are listed. Additionally, the AUROC values of each method are reported in the last row of ``AUROC”. 

\begin{table*}[!ht]
	\footnotesize
	\caption{The performance of the FRS-based method and the proposed method on detecting different types of outliers.\label{table3}}
	\begin{center}
		\setlength{\tabcolsep}{1.5mm}
		\begin{tabular}{lcccccccccccc}
			\toprule
                & \multicolumn{4}{c}{Local outliers} & \multicolumn{4}{c}{Global outliers} & \multicolumn{4}{c}{Group outliers}\\
                \cmidrule(lr){2-5} \cmidrule(lr){6-9} \cmidrule(lr){10-13}
			  & \multicolumn{2}{c}{FRS} & \multicolumn{2}{c}{Ours} & \multicolumn{2}{c}{FRS} & \multicolumn{2}{c}{Ours} & \multicolumn{2}{c}{FRS} & \multicolumn{2}{c}{Ours} \\ \cmidrule(lr){2-3} \cmidrule(lr){4-5} \cmidrule(lr){6-7} \cmidrule(lr){8-9} \cmidrule(lr){10-11} \cmidrule(lr){12-13}
         ~  & Precision & Recall & Precision & Recall & Precision & Recall & Precision & Recall & Precision & Recall & Precision & Recall\\
			\midrule
        5\% & 0.7500 & 0.2609 & 0.6250 & 0.2174 & 0.7500 & 0.2069 & 0.7500 & 0.2069 & 1.0000 & 0.2581 & 1.0000 & 0.2581 \\ 
        10\% & 0.5000 & 0.3478 & 0.6250 & 0.4348 & 0.7500 & 0.4138 & 0.8750 & 0.4828 & 0.8750 & 0.4516 & 1.0000 & 0.5161 \\ 
        15\% & 0.3333 & 0.3478 & 0.5417 & 0.5652 & 0.7083 & 0.5862 & 0.9167 & 0.7586 & 0.8333 & 0.6452 & 1.0000 & 0.7742 \\ 
        20\% & 0.3125 & 0.4348 & 0.5000 & 0.6957 & 0.5938 & 0.6552 & 0.8125 & 0.8966 & 0.8750 & 0.9032 & 0.9688 & 1.0000 \\ 
        25\% & 0.3000 & 0.5217 & 0.5000 & 0.8696 & 0.5500 & 0.7586 & 0.7000 & 0.9655 & 0.7750 & 1.0000 & 0.7750 & 1.0000 \\ 
        30\% & 0.2917 & 0.6087 & 0.4167 & 0.8696 & 0.5625 & 0.9310 & 0.6042 & 1.0000 & 0.6458 & 1.0000 & 0.6458 & 1.0000 \\ 
        35\% & 0.2857 & 0.6957 & 0.3571 & 0.8696 & 0.5179 & 1.0000 & 0.5179 & 1.0000 & 0.5536 & 1.0000 & 0.5536 & 1.0000 \\ 
        40\% & 0.2500 & 0.6957 & 0.3125 & 0.8696 & 0.4531 & 1.0000 & 0.4531 & 1.0000 & 0.4844 & 1.0000 & 0.4844 & 1.0000 \\ 
        45\% & 0.2500 & 0.7826 & 0.2778 & 0.8696 & 0.4028 & 1.0000 & 0.4028 & 1.0000 & 0.4306 & 1.0000 & 0.4306 & 1.0000 \\ 
        50\% & 0.2500 & 0.8696 & 0.2500 & 0.8696 & 0.3625 & 1.0000 & 0.3625 & 1.0000 & 0.3875 & 1.0000 & 0.3875 & 1.0000 \\ 
        55\% & 0.2386 & 0.9130 & 0.2386 & 0.9130 & 0.3295 & 1.0000 & 0.3295 & 1.0000 & 0.3523 & 1.0000 & 0.3523 & 1.0000 \\ 
        60\% & 0.2188 & 0.9130 & 0.2188 & 0.9130 & 0.3021 & 1.0000 & 0.3021 & 1.0000 & 0.3229 & 1.0000 & 0.3229 & 1.0000 \\ 
        65\% & 0.2019 & 0.9130 & 0.2019 & 0.9130 & 0.2788 & 1.0000 & 0.2788 & 1.0000 & 0.2981 & 1.0000 & 0.2981 & 1.0000 \\ 
        70\% & 0.1875 & 0.9130 & 0.1964 & 0.9565 & 0.2589 & 1.0000 & 0.2589 & 1.0000 & 0.2768 & 1.0000 & 0.2768 & 1.0000 \\ 
        75\% & 0.1750 & 0.9130 & 0.1833 & 0.9565 & 0.2417 & 1.0000 & 0.2417 & 1.0000 & 0.2583 & 1.0000 & 0.2583 & 1.0000 \\ 
        80\% & 0.1641 & 0.9130 & 0.1797 & 1.0000 & 0.2266 & 1.0000 & 0.2266 & 1.0000 & 0.2422 & 1.0000 & 0.2422 & 1.0000 \\ 
        85\% & 0.1618 & 0.9565 & 0.1691 & 1.0000 & 0.2132 & 1.0000 & 0.2132 & 1.0000 & 0.2279 & 1.0000 & 0.2279 & 1.0000 \\ 
        90\% & 0.1528 & 0.9565 & 0.1597 & 1.0000 & 0.2014 & 1.0000 & 0.2014 & 1.0000 & 0.2153 & 1.0000 & 0.2153 & 1.0000 \\ 
        95\% & 0.1513 & 1.0000 & 0.1513 & 1.0000 & 0.1908 & 1.0000 & 0.1908 & 1.0000 & 0.2039 & 1.0000 & 0.2039 & 1.0000 \\ 
        100\% & 0.1438 & 1.0000 & 0.1438 & 1.0000 & 0.1813 & 1.0000 & 0.1813 & 1.0000 & 0.1938 & 1.0000 & 0.1938 & 1.0000 \\ 
        Avg. & 0.2659 & 0.7478 & 0.3124 & 0.8391 & 0.4038 & 0.8776 & 0.4409 & 0.9155 & 0.4726 & 0.9129 & 0.4919 & 0.9274 \\
        AUROC & \multicolumn{2}{c}{0.7629}  & \multicolumn{2}{c}{0.8648} &  \multicolumn{2}{c}{0.9329}  & \multicolumn{2}{c}{0.9789} &  \multicolumn{2}{c}{0.9827}  & \multicolumn{2}{c}{1.0000}   \\ 
			\bottomrule
		\end{tabular}
	\end{center}
\end{table*}

From table \ref{table3}, it can be seen that the proposed method outperforms the FRS-based method in identifying different types of outliers. The traditional FRS-based method obtained good performance on detecting global outliers, but unsatisfactory results on detecting local and group outliers. Essentially, the traditional FRS-based method relies on the fuzzy similarity between samples to detect outliers, which is a typical distance-based outlier detection method that is good at identifying global outliers. In contrast, the proposed method takes into consideration sample similarity, density information, and multi-scale views such that local outliers can be effectively identified from low-density regions, and group outliers can be integrally recognized in coarser views. Thus, the proposed method achieved impressive performance in identifying different outliers on the synthesized datasets.

\subsection{Comparison with other state-of-the-art methods}
To evaluate the performance of the proposed method, we compared it to other state-of-the-art methods, including statistical-based methods such as Gaussian Mixture Models (GMM) \cite{Reynolds2009}, learning-based methods such as One-Class Support Vector Machines (OCSVM) \cite{Scholkopf2001}, distance-based methods such as $k$-Nearest Neighbors ($k$NN) \cite{Ramaswamy2000}, density-based methods such as Local Outlier Factor (LOF) \cite{Breunig2000}, clustering-based methods such as Cluster-Based Local Outlier Factor (CBLOF) \cite{He2003}, ensemble-based methods such as Isolation Forests (IF) \cite{Liu2008}, Feature Bagging (FB) \cite{Lazarevic2005}, Locally Selective Combination in Parallel Outlier Ensembles (LSCP) \cite{Zhao2019}, and Lightweight Online Detector of Anomalies (LODA) \cite{Pevny2016},  and FRS-based methods such as Weighted Fuzzy-Rough Density-based Anomaly detection (WFRDA) \cite{Yuan2023a} and Multi-Fuzzy Granules Anomaly Detection (MFGAD) \cite{Yuan2023b}. Each method has its own parameters that need to be tuned for better results. To make a fair comparison, we followed the settings in the work \cite{Han2022} and utilized grid search to determine the parameter values for all selected methods. To facilitate the replication of the experimental results, the optimal parameters for each method are shown in Table \ref{ps}.

\begin{table*}[!ht]
	\footnotesize
	\caption{The parameter settings of all selected methods.\label{ps}}
	\begin{center}
		\setlength{\tabcolsep}{1.5mm}
		\begin{tabular}{lcccccccccccc}
			\toprule
                    \multirow{-2}{*}{~}  & $k$NN	& IF &	OCSVM &	LOF & FB	& LSCP	& LODA &	CBLOF &	WFRDA &	MFGAD & \multicolumn{2}{c}{Ours}   \\
                    \cmidrule(lr){2-13}
                    ~ & $k_\text{kNN}$ & $n_\text{IF}$ & $ker$ &  $n_\text{LOF}$ & $n_\text{FB}$ & $n_\text{LSCP}$ & $n_\text{LODA}$ & $k_\text{CBLOF}$ & $\delta_\text{WFRDA}$ & $\delta_\text{MFGAD}$  & $\lambda$ & $\delta$    \\
                    \midrule
                    Arrhyth & 20 & 100 & rbf & 50 & 5 & 3 & 20 & 6 & 2.0 & 0.2 & 100 & 1.6 \\ 
                    Autos & 20 & 100 & rbf & 50 & 5 & 3 & 20 & 10 & 0.2 & 2.0 & 100 & 1.6 \\ 
                    Breast & 20 & 100 & rbf & 5 & 3 & 3 & 10 & 8 & 0.2 & 0.2 & 10 & 0.1 \\ 
                    Cardio & 20 & 50 & rbf & 50 & 3 & 3 & 20 & 8 & 0.8 & 0.8 & 50 & 1.0 \\ 
                    Cardioto & 20 & 100 & rbf & 50 & 10 & 3 & 5 & 6 & 0.2 & 0.2 & 50 & 0.1 \\ 
                    Chess & 10 & 50 & rbf & 50 & 10 & 3 & 5 & 10 & 0.6 & 0.2 & 10 & 0.4 \\ 
                    Hepat & 20 & 100 & poly & 50 & 20 & 3 & 20 & 8 & 0.2 & 0.2 & 10 & 1.3 \\ 
                    Iono & 3 & 100 & rbf & 5 & 5 & 3 & 20 & 8 & 1.0 & 1.4 & 10 & 0.1 \\ 
                    Iris & 5 & 50 & rbf & 50 & 10 & 3 & 20 & 8 & 0.2 & 1.4 & 10 & 0.1 \\ 
                    Mammo & 20 & 100 & rbf & 50 & 10 & 3 & 10 & 10 & 0.2 & 0.2 & 100 & 0.1 \\ 
                    Carpet & 5 & 10 & sigmoid & 20 & 5 & 3 & 15 & 8 & 0.2 & 0.2 & 100 & 1.9 \\ 
                    Metal & 5 & 10 & linear & 20 & 5 & 3 & 15 & 8 & 0.2 & 0.2 & 50 & 1.6 \\ 
                    Pill & 3 & 100 & linear & 20 & 20 & 3 & 15 & 10 & 0.2 & 0.2 & 100 & 1.3 \\ 
                    Pen & 20 & 100 & rbf & 5 & 10 & 3 & 15 & 8 & 0.6 & 2.0 & 50 & 0.7 \\ 
                    Sat & 20 & 10 & rbf & 10 & 5 & 3 & 20 & 10 & 0.2 & 0.2 & 10 & 1.3 \\ 
                    Spam & 20 & 100 & rbf & 5 & 5 & 3 & 5 & 8 & 0.2 & 0.2 & 10 & 1.6 \\ 
                    Thyroid & 20 & 10 & rbf & 20 & 20 & 3 & 10 & 10 & 0.4 & 1.2 & 50 & 1.0 \\ 
                    WDBC & 20 & 100 & rbf & 50 & 10 & 3 & 5 & 10 & 0.4 & 1.6 & 10 & 0.4 \\ 
                    Wine & 20 & 100 & rbf & 50 & 5 & 3 & 20 & 6 & 2.0 & 0.2 & 10 & 0.1 \\ 
                    WPBC & 10 & 100 & rbf & 20 & 3 & 3 & 20 & 6 & 0.2 & 2.0 & 100 & 1.0 \\ 
			\bottomrule
		\end{tabular}
	\end{center}
\end{table*}

With the parameter settings in Table \ref{ps}, the performance and the rank information of the proposed method and other methods on all selected datasets are presented in Table \ref{performance}, where the values in each cell denote the AUROC and rank of each method, respectively, and the best performance on each dataset is boldfaced. Note that the methods such as IF, FB, LSCP, and CBLOF, were initialized randomly. Thus, these methods were repeated 5 times, and their average performance was recorded.

\begin{table*}[!ht]
	\footnotesize
	\caption{The performance of the selected methods over all datasets. \label{performance}}
	\begin{center}
		\setlength{\tabcolsep}{1.0mm}
            \resizebox{\linewidth}{!}{ 
		\begin{tabular}{lcccccccccccc}
			\toprule
            ~ & $k$NN & GMM & IF & OCSVM & LOF & FB & LSCP & LODA & CBLOF & WFRDA & MFGAD & Ours \\ 
                \midrule
                    Arrhyth & 0.8041 (3) & 0.7585 (10) & 0.8056 (2) & 0.7872 (7) & 0.7889 (5) & 0.7639 (9) & 0.7581 (11) & 0.7839 (8) & 0.7882 (6) & 0.7896 (4) & 0.6771 (12) & \textbf{0.8630} (1)\\
                    Autos & 0.6022 (6) & 0.5247 (10) & 0.6344 (2) & 0.5847 (7) & 0.6024 (5) & 0.5216 (11) & 0.5058 (12) & 0.5767 (9) & 0.5802 (8) & 0.6249 (3) & 0.6047 (4) & \textbf{0.8322} (1)\\
                    Breast & 0.9819 (6) & 0.9724 (7) & 0.9875 (3) & 0.8052 (9) & 0.4537 (10) & 0.3337 (12) & 0.3883 (11) & 0.9918 (2) & 0.9688 (8) & 0.9855 (4) & 0.9838 (5) & \textbf{0.9960} (1)\\
                    Cardio & 0.8499 (9) & 0.8965 (7) & 0.9298 (3) & 0.9286 (4) & 0.7468 (10) & 0.6493 (11) & 0.6305 (12) & 0.9422 (2) & 0.8914 (8) & 0.9217 (5) & 0.8967 (6) & \textbf{0.9606} (1)\\
                    Cardioto & 0.5794 (9) & 0.6025 (6) & 0.7113 (4) & 0.7872 (2) & 0.5946 (7) & 0.5859 (8) & 0.5673 (10) & 0.7606 (3) & 0.6689 (5) & 0.5438 (11) & 0.5243 (12) & \textbf{0.8631} (1)\\
                    Chess & 0.8908 (3) & 0.8649 (8) & 0.8894 (4) & 0.8675 (6) & 0.9031 (2) & 0.7291 (11) & 0.8673 (7) & 0.6153 (12) & 0.8388 (9) & 0.7722 (10) & 0.8740 (5) & \textbf{0.9278} (1)\\
                    Hepat & 0.7658 (4) & 0.6739 (10) & 0.7132 (7) & 0.7061 (8) & 0.7968 (3) & 0.6900 (9) & 0.6257 (12) & 0.7139 (6) & 0.6482 (11) & 0.8025 (2) & 0.7302 (5) & \textbf{0.8553} (1)\\
                    Iono & \textbf{1.0000} (3) & \textbf{1.0000} (3) & \textbf{1.0000} (3) & 0.9991 (6) & 0.9911 (11) & 0.9925 (10) & 0.9907 (12) & 0.9985 (8) & \textbf{1.0000} (3) & 0.9926 (9) & 0.9987 (7) & \textbf{1.0000} (3)\\
                    Iris & 0.9900 (7) & 0.9455 (11) & 0.9795 (9) & \textbf{1.0000} (2) & 0.9982 (5) & 0.9805
                    (8) & 0.9518 (10) & 0.9311 (12) & 0.9993 (4) & \textbf{1.0000} (2) & 0.9973 (6) & \textbf{1.0000} (2)\\
                    Mammo & 0.8461 (5) & 0.8602 (4) & 0.8671 (3) & 0.8412 (6) & 0.8213 (8) & 0.7809 (11) & 0.7332 (12) & 0.8682 (2) & 0.8302 (7) & 0.7810 (10) & 0.7921 (9) & \textbf{0.8784} (1)\\
                    Carpet & 0.7248 (3) & 0.7147 (9) & 0.7027 (10) & 0.7268 (2) & 0.7180 (7) & 0.7217 (4) & 0.7160 (8) & 0.6981 (11) & 0.7213 (5) & 0.7183 (6) & 0.6494 (12) & \textbf{0.8405} (1)\\
                    Metal & 0.7159 (4) & \textbf{0.7404} (1) & 0.6483 (9) & 0.6466 (10) & 0.7097 (6) & 0.7165 (3) & 0.7105 (5) & 0.6636 (8) & 0.6994 (7) & 0.4563 (12) & 0.6177 (11) & 0.7348 (2)\\
                    Pill & 0.6771 (3) & \textbf{0.7033} (1) & 0.6387 (9) & 0.6224 (11) & 0.6688 (5) & 0.6696 (4) & 0.6685 (6) & 0.6449 (8) & 0.6573 (7) & 0.6234 (10) & 0.5903 (12) & 0.7006 (2)\\
                    Pen & 0.7677 (8) & 0.7628 (9) & 0.9620 (3) & 0.9354 (7) & 0.5437 (10) & 0.4844 (11) & 0.4794 (12) & 0.9608 (4) & \textbf{0.9724} (1) & 0.9423 (5) & 0.9414 (6) & 0.9696 (2)\\
                    Sat & 0.9745 (7) & 0.9838 (5) & 0.9939 (3) & 0.9747 (6) & 0.5949 (10) & 0.5495 (11) & 0.5369 (12) & 0.9914 (4) & \textbf{0.9989} (1) & 0.9594 (9) & 0.9732 (8) & 0.9943 (2)\\
                    Spam & 0.5618 (5) & 0.5295 (7) & 0.6587 (3) & 0.5251 (8) & 0.4726 (10) & 0.4459 (12) & 0.4599 (11) & 0.5170 (9) & 0.5592 (6) & 0.7444 (2) & 0.6235 (4) & \textbf{0.7884} (1)\\
                    Thyroid & 0.6566 (5) & 0.6716 (3) & 0.6640 (4) & 0.6539 (6) & 0.5331 (10) & 0.5655 (8) & 0.5334 (9) & 0.6405 (7) & 0.6794 (2) & 0.5307 (11) & 0.3793 (12) & \textbf{0.6818} (1)\\
                    WDBC & 0.9813 (7) & 0.9603 (8) & 0.9886 (6) & 0.9935 (4) & 0.9517 (9) & 0.4830 (11) & 0.4356 (12) & 0.9902 (5) & 0.8970 (10) & \textbf{0.9988} (1) & 0.9961 (3) & 0.9971 (2)\\
                    Wine & 0.9261 (3) & 0.6496 (12) & 0.8024 (8) & 0.6941 (11) & 0.9202 (4) & 0.8953 (6) & 0.8748 (7) & 0.9008 (5) & 0.9585 (2) & 0.7807 (9) & 0.7626 (10) & \textbf{0.9992} (1)\\
                    WPBC & 0.5323 (3) & 0.4739 (12) & 0.5138 (9) & 0.4743 (11) & 0.5184 (8) & 0.5244 (5) & 0.5185 (7) & 0.5295 (4) & 0.5061 (10) & 0.5222 (6) & 0.5669 (2) & \textbf{0.5722} (1)\\
                    Avg. & 0.7914 (5.15) & 0.7645 (7.15) & 0.8045 (5.20) & 0.7777 (6.65) & 0.7164 (7.25) & 0.6542 (8.75) & 0.6476 (9.90) & 0.7860 (6.45) & 0.7932 (6.00) & 0.7745 (6.55) & 0.7590 (7.55) & \textbf{0.8727} (1.40)\\
			\bottomrule
		\end{tabular}}
	\end{center}
\end{table*}

Table \ref{performance} shows that the selected methods are capable of detecting outliers in datasets, but their performance varies greatly. Specifically, the distance-based methods such as $k$NN, WFRDA, and MFGAD, obtained good results on most datasets. The reason for this may be that distance-based methods are good at detecting global outliers, while outlier datasets are generally generated from multi-class data by removing samples from minority classes, thereby resulting in more global and group outliers. The clustering-based methods such as CBLOF also achieved satisfactory performance, but slightly different from distance-based methods. For example, the CBLOF performs well on the datasets of ``Pen” and ``stai”, but the $k$NN fails. One possible explanation is that distance-based methods excel at detecting global outliers but face challenges in identifying group outliers, while clustering-based methods exhibit the opposite characteristic. The performance of GMM and OCSVM is inferior to that of CBLOF and $k$NN. This may be due to the fact that these methods make the assumption for the distribution of samples, while this assumption may not be met on some datasets. For example, on the datasets of ``Hepa” and ``Wine”, the results of GMM and OCSVM are both unsatisfactory. The density-based methods such as LOF and the methods using LOF as the base detector such as FB and LSCP gained poor performance on most datasets. The reason may be attributed to that these methods only rely on the density information to detect outliers, which is ineffective in identifying global and group outliers. 
Notably, ensemble-based methods, particularly IF, attained promising results. The reason for this lies in the fact that ensemble learning is beneficial for improving performance by diversifying base detectors and fusing results from different perspectives.

By integrating fuzzy similarity and density information of samples, the proposed method can detect global and local outliers effectively, while the multi-scale granular balls improve the capability of recognizing group outliers. Moreover, a weighted SVM is trained to further refine the results from multi-scale views. Thus, the proposed method can achieve very satisfactory performance in detecting different types of outliers. 
By averaging the results on all selected datasets, the proposed method improves over single view methods such as $k$NN, GMM, OCSVM, LOF, CBLOF, WFRDA, and MFGAD by 10.27\%, 14.16\%, 12.22\%, 21.82\%, 10.03\%, 12.68\%, and 14.99\%, respectively, and over ensemble-based methods such as IF, FB, LSCP, and LODA by 8.47\%, 33.41\%, 34.76\%, and 11.04\%, respectively.

\subsection{Statistical significance analysis}
Statistical significance analysis was conducted to further compare the proposed method with other selected methods. The Friedman test was first used to examine whether there is a statistically significant difference in performance among all methods. In the Friedman test, the statistical variable $\tau_F$ is defined as \cite{Demšar2006}
\begin{equation}
    \begin{aligned}
        \tau_F=\frac{(N-1)\tau_{\chi^2}}{N(M-1)-\tau_{\chi^2}},
    \end{aligned}
\end{equation}

\begin{equation}
    \begin{aligned}
        \tau_{\chi^2}=\frac{12N}{M(M+1)}{\big(\sum_{i=1}^Mr_i^2-\frac{M(M+1)^2}{4}\big),}
    \end{aligned}
\end{equation}
where $N$ and $M$ denote the number of datasets and methods, respectively, and $r_i$ stands for the average rank value of the $i$-th method on all datasets. 

The variable $\tau_F$ follows an $F$-distribution with degrees of freedom $M-1$ and $(M-1)(N-1)$. In our experiments, there are $M=12$ methods and $N=20$ datasets, and the critical value of $F(11,209)$ at the significance level of $\phi = 0.05$ is $1.8346$. According to the performance rank results in Table \ref{performance}, the calculated value of $\tau_F$ is $9.7544$, which is significantly greater than the critical value of $F(11,209)=1.8346$. The Friedman test reveals that the performance difference of all selected methods is statistically significant.

To further examine the difference between each pair of the selected methods in performance, the post hoc Nemenyi test was performed. The critical value in the Nemenyi test is defined as \cite{Demšar2006}
\begin{equation}
    \begin{aligned} 
        CD_{\phi}=q_{\phi}\sqrt{\frac{M(M+1)}{6N}},
    \end{aligned}
\end{equation}
where $q_\phi$ is the critical value of the Tukey distribution at a significance level of $\phi$. 

\begin{figure}[!ht]
	\begin{centering}              
            \includegraphics[scale=0.048]
            {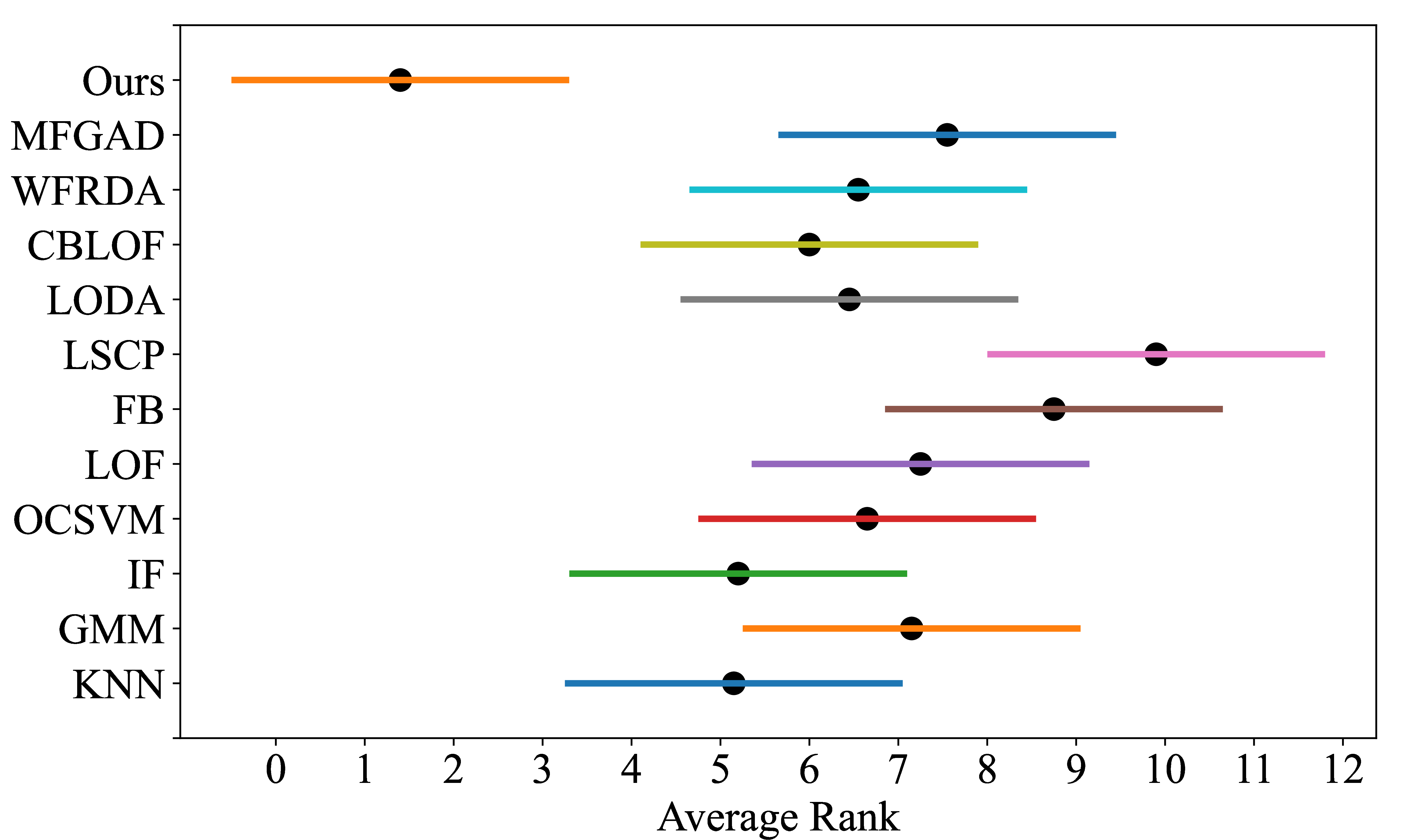}
        \end{centering}
	\caption{The Friedman test diagram.}\label{f7}
\end{figure}

At the significance level of $\phi=0.05$, the critical value for the Tukey distribution is $q_{0.05}=3.2680$ $(M=12)$ and the critical value of the Nemenyi test is $CD_{0.05} = 3.7261$ $(M=12, N=20)$. The final diagram of the Friedman test is shown in Fig. \ref{f7}.

In Fig. \ref{f7}, each dot denotes the average rank value of the corresponding method, and the line centered on the dot indicates the size of the critical value in the Nemenyi test, i.e., $CD_{0.05} = 3.7261$. If the difference between the rank values of two methods is greater than $3.7261$, there is a significant difference between the two methods in performance; otherwise, there is no significant difference. Fig. \ref{f7} shows that the proposed method is statistically significantly better than other methods. Among all selected methods, LSCP has the lowest average rank value and is statistically significantly inferior to the proposed method, CBLOF, IF, and $k$NN. In addition, the significant differences between other methods are not tested, although their performance on the selected datasets is different from each other.

\subsection{Ablation experiments}
To evaluate the impact of the density information, multi-scale granular balls, and weighted SVM on the performance of the proposed method, we conducted ablation experiments on all selected datasets, and the results are shown in Table \ref{ablation}, where each column represents the AUROC values of a corresponding method on different datasets, and the row of ``Avg.” denotes the average performance over all datasets.

\begin{table*}[!ht]
	\footnotesize
	\caption{The results of the ablation experiments. \label{ablation}}
	\begin{center}
		\setlength{\tabcolsep}{1.0mm}
		\begin{tabular}{lcccccccc}
			\toprule
        Fuzzy granule density &  \ding{56} & \ding{52} & \ding{56} & \ding{56} & \ding{56} & \ding{52} & \ding{52} & \ding{52} \\ 
	Multi-scale granular balls & \ding{56} & \ding{56} & \ding{52} & \ding{56} & \ding{52} & \ding{56} & \ding{52} & \ding{52}	\\
        Weighted SVM & \ding{56} & \ding{56} & \ding{56} & \ding{52} & \ding{52} & \ding{52} & \ding{56} & \ding{52}\\
 \midrule
Arrhyth & 0.6326 & 0.5941 & 0.8061 & 0.6466 & 0.8473 & 0.8467 & 0.7765 & 0.8630 \\
Autos & 0.5881 & 0.5923 & 0.7846 & 0.7413 & 0.7933 & 0.6642 & 0.7434 & 0.8322 \\
Breast & 0.9400 & 0.9400 & 0.9112 & 0.9771 & 0.9912 & 0.9771 & 0.9706 & 0.9960 \\
Cardio & 0.7644 & 0.7567 & 0.9125 & 0.8071 & 0.9496 & 0.8160 & 0.9000 & 0.9606 \\
Cardioto & 0.3754 & 0.4140 & 0.7863 & 0.3400 & 0.8562 & 0.3775 & 0.8080 & 0.8631 \\
Chess & 0.5943 & 0.5943 & 0.8818 & 0.6393 & 0.9036 & 0.6393 & 0.8942 & 0.9278 \\
Hepat & 0.6223 & 0.5924 & 0.5408 & 0.6958 & 0.6498 & 0.7095 & 0.6900 & 0.8553 \\
Iono & 0.9963 & 0.9981 & 0.9828 & 0.9957 & 1.0000 & 0.9969 & 0.9891 & 1.0000 \\
Iris & 1.0000 & 0.9782 & 0.9982 & 1.0000 & 1.0000 & 1.0000 & 0.9955 & 1.0000 \\
Mammo & 0.8474 & 0.8624 & 0.8082 & 0.7568 & 0.7025 & 0.8364 & 0.8552 & 0.8784 \\
Carpet & 0.6060 & 0.6435 & 0.6934 & 0.6990 & 0.8221 & 0.8360 & 0.6773 & 0.8405 \\
Metal & 0.6476 & 0.5283 & 0.6525 & 0.5845 & 0.7224 & 0.5657 & 0.6507 & 0.7348 \\
Pill & 0.5821 & 0.5399 & 0.6146 & 0.6140 & 0.6776 & 0.4728 & 0.5956 & 0.7006 \\
Pen & 0.7522 & 0.8873 & 0.9005 & 0.7657 & 0.9066 & 0.9152 & 0.8897 & 0.9696 \\
Sat & 0.5204 & 0.5204 & 0.9918 & 0.8430 & 0.9943 & 0.8430 & 0.9920 & 0.9943 \\
Spam & 0.6627 & 0.5716 & 0.6742 & 0.6340 & 0.7459 & 0.6756 & 0.6865 & 0.7884 \\
Thyroid & 0.5095 & 0.5500 & 0.5929 & 0.4644 & 0.6355 & 0.5283 & 0.6602 & 0.6818 \\
WDBC & 0.9930 & 0.9952 & 0.9845 & 0.9979 & 0.9941 & 0.9973 & 0.9815 & 0.9971 \\
Wine & 0.6328 & 0.7504 & 0.9202 & 0.2328 & 0.9992 & 0.9983 & 0.9790 & 0.9992 \\
WPBC & 0.4258 & 0.5525 & 0.4616 & 0.3671 & 0.4870 & 0.6156 & 0.5597 & 0.5722 \\
Avg. & 0.6846 & 0.6931 & 0.7949 & 0.6901 & 0.8339 & 0.7656 & 0.8147 & 0.8727 \\
			\bottomrule
		\end{tabular}
	\end{center}
\end{table*}

From Table \ref{ablation}, the following observations can be made:
\begin{itemize}
    \item Each component of the proposed method has a positive effect on the performance of the FRS-based method. Among the three components, the strategy of multi-scale granular balls greatly enhances the detection of group outliers and brings the greatest improvement in performance, with an increase of $16.11\%$ over the FRS-based method.
    \item A good fusion strategy can significantly improve the performance of the multi-scale outlier detection method. The weight SVM is trained on reliable outliers and inliers determined by the three-way decision and can be used to refine outlier detection results fused from different views. The multi-scale outlier detection using the trained SVM achieves a performance improvement of $4.91\%$.
    \item Density information is essentially complementary to sample distance when evaluating sample relationships and can be of great benefit to distance-based outlier detection methods. Relative fuzzy granule density captures the distribution of samples in local regions and has a substantive effect on improving the capability of detecting local outliers. By introducing density information, the performance of the multi-scale outlier detection method is further improved by $4.65\%$.
\end{itemize}

\subsection{Effect of the number of views on performance}

\begin{figure*}[!ht]
	\footnotesize
	\begin{centering}
            \subfigure[]{\includegraphics[scale=0.023]{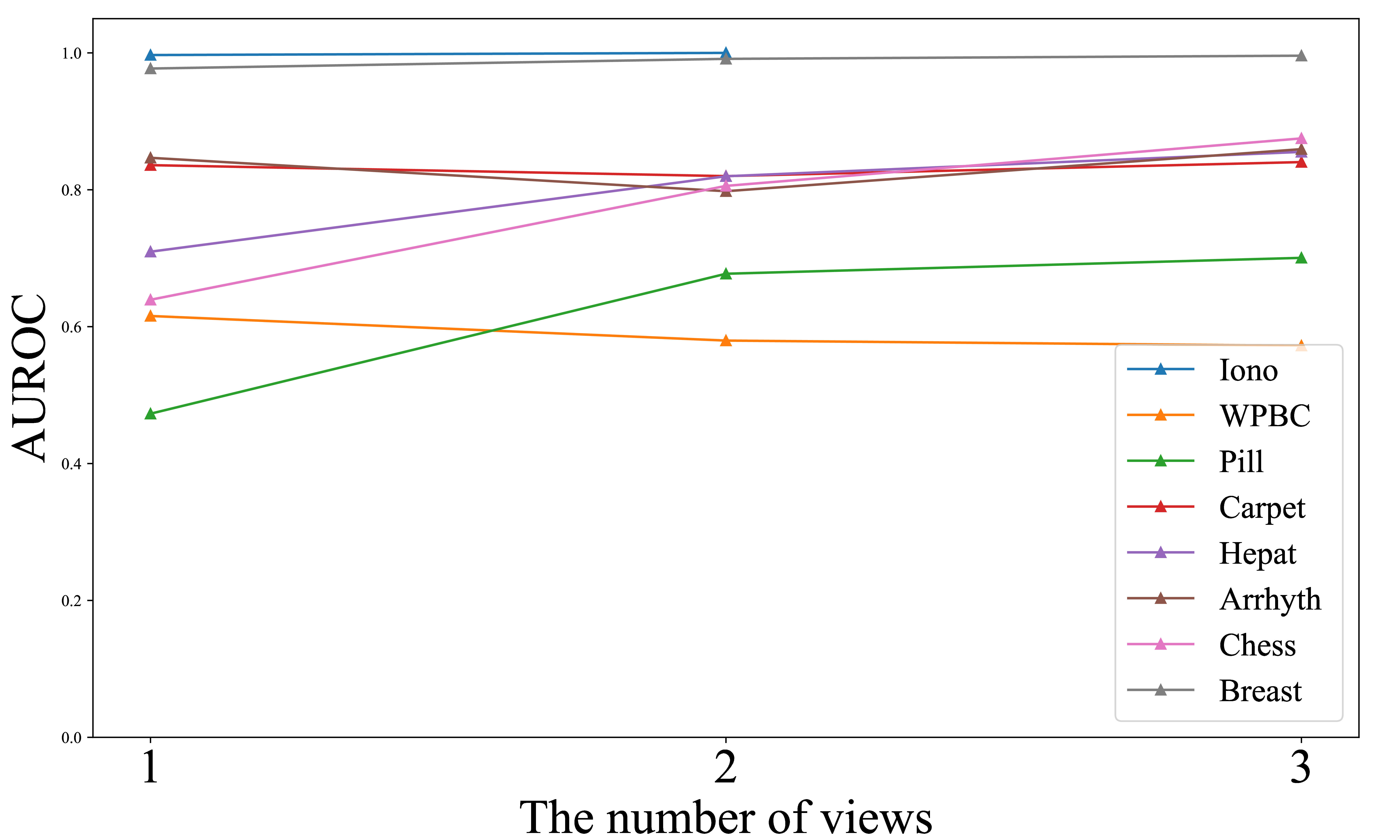}}
		\subfigure[]{\includegraphics[scale=0.023]{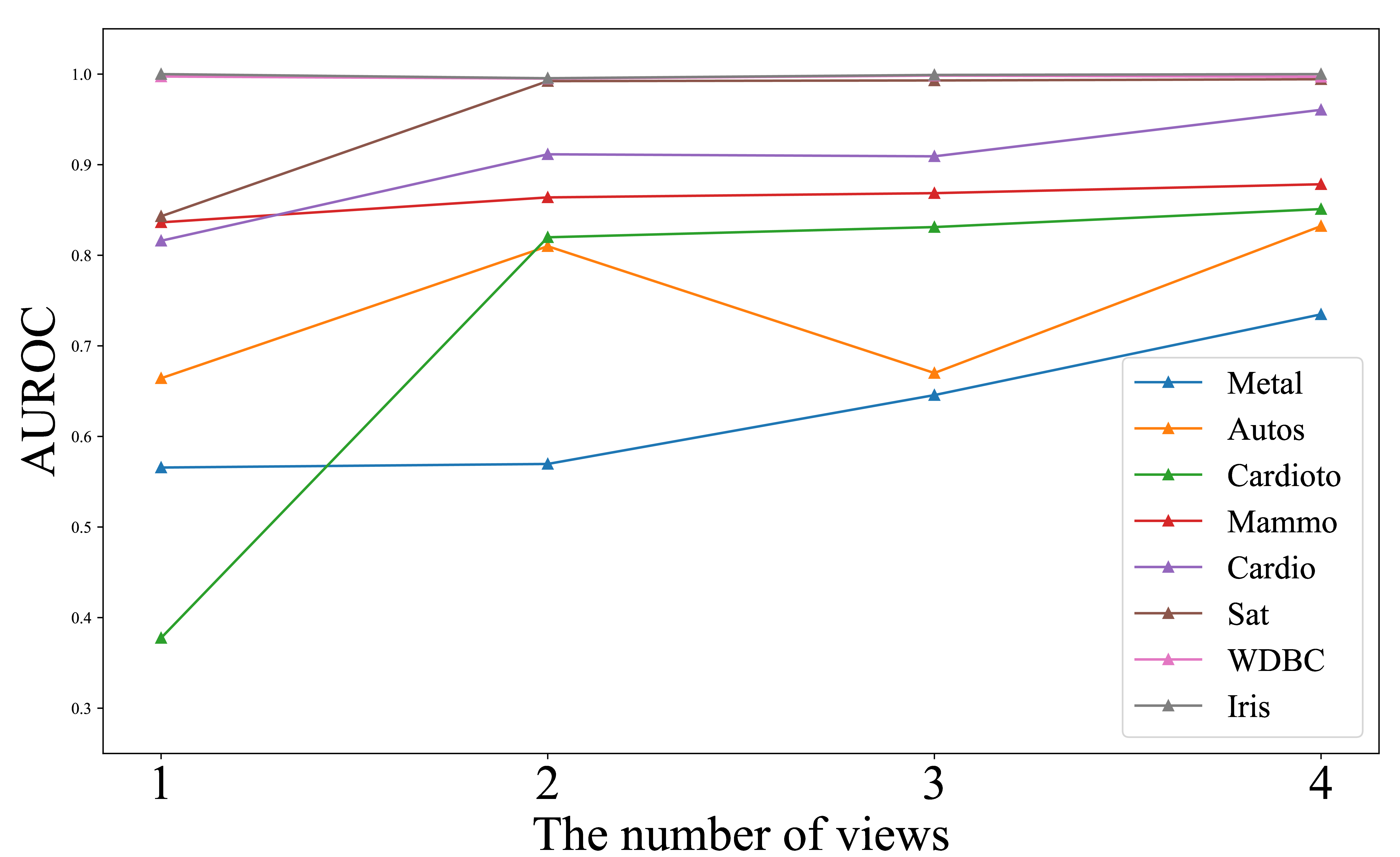}}
		\subfigure[]{\includegraphics[scale=0.023]{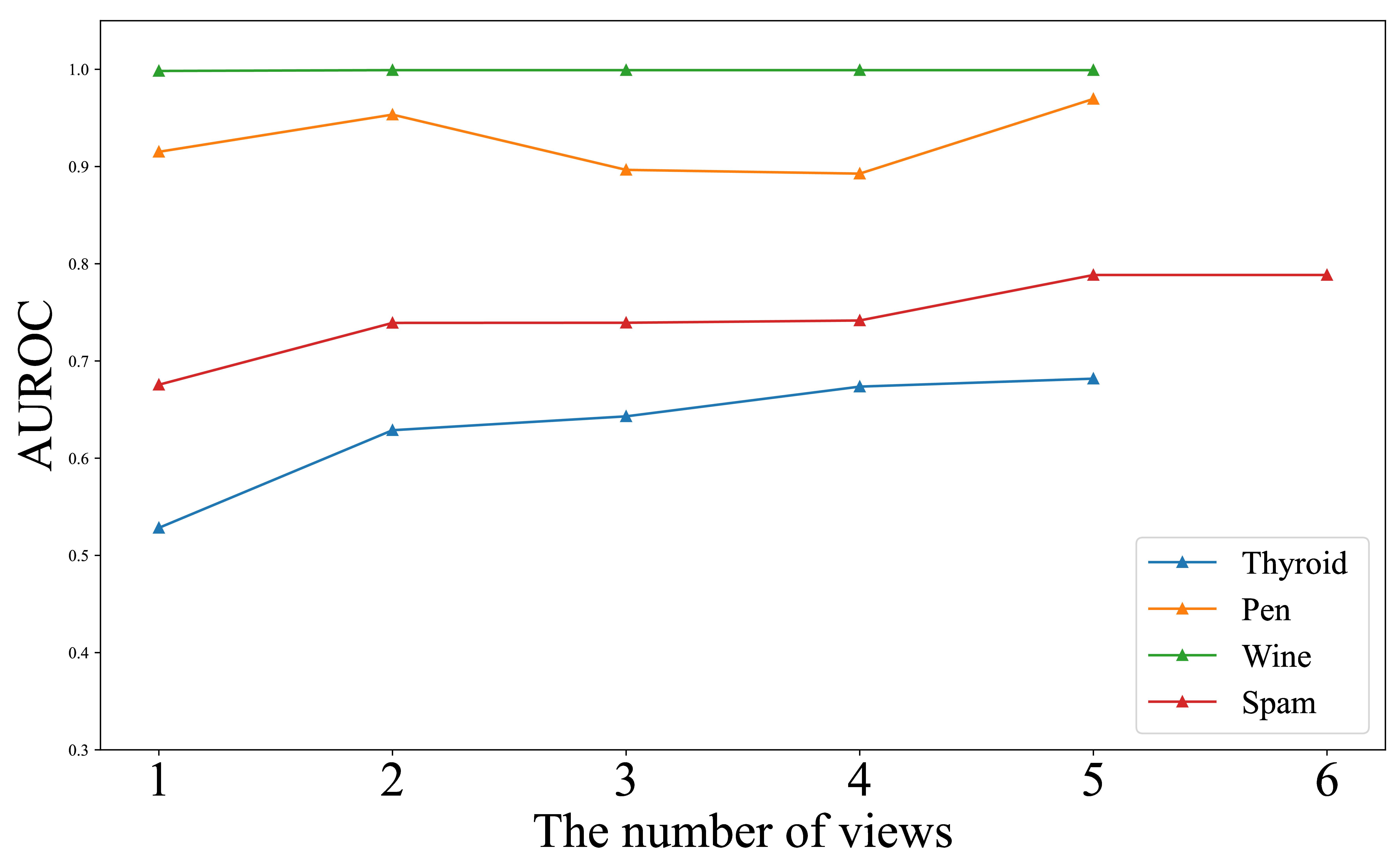}}
		\par\end{centering}
	\caption{The performance with different number of views. (a) the datasets with 2-3 views. (b) the datasets with 4 views. (c) the datasets with 5-6 views. Note that the number of views is adaptively determined by the granular ball generation method.}
        \label{view number2}
\end{figure*}

To examine the impact of the number of views on the performance, the proposed method with different number of views is performed, and the results are shown in Fig. \ref{view number2}.

As shown in Fig. \ref{view number2}, the number of views varies across different datasets, and the performance of the proposed method increases when more views are added. Additionally, the performance tends to be stable when the number of views is approaching the maximum value. These results demonstrate that the introduction of multi-scale views is very beneficial for outlier detection.

\subsection{Analysis of time complexity}
To examine the efficiency of the proposed method, we performed a comparative analysis of the computational complexity, and the results are shown in Table \ref{time complex}, where $T$ denotes the number of iterations, and $k_{i}$ represents the number of base models used in the ensemble-based method $i$.

\begin{table*}[!ht]
\footnotesize
\caption{The comparison of the selected methods in time complexity.}
\label{time complex}
\begin{center}
\setlength{\tabcolsep}{1.0mm}
    \resizebox{\linewidth}{!}{ 
    \begin{tabular}{lcccccccccccc}
		\toprule
        $k$NN & GMM & IF & OCSVM & LOF & FB & LSCP & LODA & CBLOF & WFRDA & MFGAD & Ours \\
        \midrule
        $O(|A||U|^2)$ & $O(T|A||U|^2)$ & $O(k_{IF}|A||U|\log|U|)$ & $O(|A||U|^3)$ & $O(|A||U|^2)$ & $O(k_{FB} |A||U|^2)$ & $O(k_{LSCP} |A||U|^2)$ & $O(|A|^{-\frac{1}{2}}|U|)$ & $O(T \times k|A||U|)$ & $O(|A||U|^2)$ & $O(|A||U|^2)$ & $O(|A||U|^2 \log |U|)$  \\
		\bottomrule
    \end{tabular}}
\end{center}
\end{table*}

From Table \ref{time complex}, it can be seen that due to the utilization of multiple views, the time complexity of the proposed method is slightly larger than that of single model methods, such as $k$NN, GMM, LOF, CBLOF, WFRDA, and MFGAD. However, as shown in Fig. \ref{view number2}, the number of views is generally small, and the gap between these methods is thus limited. Moreover, there is no significant difference in time complexity between the proposed method and most ensemble-based methods. Although the time complexity of IF is relatively low in theory, it requires the use of nearly 50-100 trees to achieve satisfactory performance on most datasets.
\color{black}

\section{Conclusions}\label{section 5}
In real-world applications, available data may contain different types of outliers, including local outliers, global outliers, and group outliers, posing a great challenge to outlier detection methods. In this study, we introduced the concept of relative fuzzy granule density to improve the ability of fuzzy rough sets-based methods for identifying local outliers and developed a multi-scale granular ball generation method to enhance the separability of group outliers. To effectively fuse the results obtained from multi-scale views, a weight SVM is trained on the reliable outliers and inliers acquired by the three-way decision to generate an accurate outlier probability vector. Extensive comparative experiments and their statistical significance analysis demonstrated that the proposed method outperforms other state-of-the-art methods in identifying different types of outliers. Exploring a more effective fuse strategy for multi-scale outlier detection is worthy of further investigation, and fuzzy rough sets-based outlier detection for semi-supervised data will be our future research work.

\section{Acknowledgments}
We would like to thank the editor-in-chief, editor, and anonymous reviewers for their insightful and constructive comments to greatly improve the quality of the paper. This work was supported in part by the National Natural Science Foundation of China (No. 62476171, 62476172, 62076164), Guangdong Basic and Applied Basic Research Foundation (No. 2024A1515011367), Guangdong Provincial Key Laboratory (No. 2023B1212060076), and Shenzhen Institute of Artificial Intelligence and Robotics for Society.

\bibliographystyle{IEEEtran}
\bibliography{main}

\begin{thebibliography}{10}
\providecommand{\url}[1]{#1}
\csname url@samestyle\endcsname
\providecommand{\newblock}{\relax}
\providecommand{\bibinfo}[2]{#2}
\providecommand{\BIBentrySTDinterwordspacing}{\spaceskip=0pt\relax}
\providecommand{\BIBentryALTinterwordstretchfactor}{4}
\providecommand{\BIBentryALTinterwordspacing}{\spaceskip=\fontdimen2\font plus
\BIBentryALTinterwordstretchfactor\fontdimen3\font minus \fontdimen4\font\relax}
\providecommand{\BIBforeignlanguage}[2]{{%
\expandafter\ifx\csname l@#1\endcsname\relax
\typeout{** WARNING: IEEEtran.bst: No hyphenation pattern has been}%
\typeout{** loaded for the language `#1'. Using the pattern for}%
\typeout{** the default language instead.}%
\else
\language=\csname l@#1\endcsname
\fi
#2}}
\providecommand{\BIBdecl}{\relax}
\BIBdecl

\bibitem{Boukerche2021}
A.~Boukerche, L.~Zheng, and O.~Alfandi, ``Outlier detection: Methods, models, and classification,'' \emph{ACM Comput. Surv.}, vol.~53, no.~3, p. Article 55, 2020.

\bibitem{Ma2021}
X.~Ma, J.~Wu, S.~Xue, J.~Yang, Q.~Z. Sheng, and H.~Xiong, ``A comprehensive survey on graph anomaly detection with deep learning,'' \emph{IEEE Trans. Knowl. Data Eng.}, vol. Early Access, Oct. 2021.

\bibitem{Pang2022}
G.~Pang, C.~Shen, L.~Cao, and A.~V.~D. Hengel, ``Deep learning for anomaly detection: A review,'' \emph{ACM Comput. Surv.}, vol.~54, no.~2, pp. 1--38, 2021.

\bibitem{WangYZ2023}
Y.~Wang, Z.~Yu, and L.~Zhu, ``Intrusion detection for high-speed railways based on unsupervised anomaly detection models,'' \emph{Appl. Intell.}, vol.~53, no.~7, pp. 8453--8466, 2022.

\bibitem{Chen2022}
T.~Chen and C.~Tsourakakis, ``Antibenford subgraphs: Unsupervised anomaly detection in financial networks,'' in \emph{Proc. ACM SIGKDD Int’l Conf. Knowledge Discovery and Data Mining}, 2022, pp. 2762--2770.

\bibitem{Zhou2019}
J.~T. Zhou, J.~Du, H.~Zhu, X.~Peng, Y.~Liu, and R.~S.~M. Goh, ``Anomalynet: An anomaly detection network for video surveillance,'' \emph{IEEE Trans. Inf. Forensics Secur.}, vol.~14, no.~10, pp. 2537--2550, 2019.

\bibitem{Reynolds2009}
G.~Steinbuss and K.~Böhm, ``Benchmarking unsupervised outlier detection with realistic synthetic data,'' \emph{ACM Trans. Knowl. Discov. Data}, vol.~15, no.~4, pp. 1--20, 2021.

\bibitem{Angiulli2002}
F.~Angiulli and C.~Pizzuti, ``Fast outlier detection in high dimensional spaces,'' in \emph{Principles of Data Mining and Knowledge Discovery}.\hskip 1em plus 0.5em minus 0.4em\relax Springer, 2002, pp. 15--27.

\bibitem{Breunig2000}
M.~M. Breunig, H.-P. Kriegel, R.~T. Ng, and J.~Sander, ``Lof: Identifying density-based local outliers,'' in \emph{Proc. ACM SIGMOD Int. Conf. on Management of Data}, 2000, pp. 93--104.

\bibitem{He2003}
Z.~He, X.~Xu, and S.~Deng, ``Discovering cluster-based local outliers,'' \emph{Pattern Recognit. Lett.}, vol.~24, pp. 1641--1650, 2003.

\bibitem{Lazarevic2005}
A.~Lazarevic and V.~Kumar, ``Feature bagging for outlier detection,'' in \emph{Proc. Eleventh ACM SIGKDD Int. Conf. on Knowledge Discovery in Data Mining}, 2005, pp. 157--166.

\bibitem{Ramaswamy2000}
S.~Ramaswamy, R.~Rastogi, and K.~Shim, ``Efficient algorithms for mining outliers from large data sets,'' in \emph{Proc. ACM SIGMOD Int. Conf. on Management of Data}, 2000, pp. 427--438.

\bibitem{Pawlak1982}
Z.~Pawlak, J.~Grzymala-Busse, R.~Slowinski, and W.~Ziarko, ``Rough sets,'' \emph{Commun. ACM}, vol.~38, no.~11, pp. 88--95, 1995.

\bibitem{XiaBW2023}
S.~Xia, S.~Wu, X.~Chen, G.~Wang, X.~Gao, Q.~Zhang, E.~Giem, and Z.~Chen, ``An efficient and accurate rough set for feature selection, classification, and knowledge representation,'' \emph{IEEE Trans. Knowl. Data Eng.}, vol.~35, no.~8, pp. 7724--7735, 2023.

\bibitem{Jiang2009}
F.~Jiang, Y.~Sui, and C.~Cao, ``Some issues about outlier detection in rough set theory,'' \emph{Expert Syst. Appl.}, vol.~36, no.~3, pp. 4680--4687, 2009.

\bibitem{Jiang2008}
------, ``A rough set approach to outlier detection,'' \emph{Int. J. Gen. Syst.}, vol.~37, no.~5, pp. 519--536, 2008.

\bibitem{Jiang2011}
------, ``A hybrid approach to outlier detection based on boundary region,'' \emph{Pattern Recognit. Lett.}, vol.~32, no.~14, pp. 1860--1870, 2011.

\bibitem{Albanese2014}
A.~Albanese, S.~K. Pal, and A.~Petrosino, ``Rough sets, kernel set, and spatiotemporal outlier detection,'' \emph{IEEE Trans. Knowl. Data Eng.}, vol.~26, no.~1, pp. 194--207, 2014.

\bibitem{Jiang2010}
F.~Jiang, Y.~Sui, and C.~Cao, ``An information entropy-based approach to outlier detection in rough sets,'' \emph{Expert Syst. Appl.}, vol.~37, no.~9, pp. 6338--6344, 2010.

\bibitem{Jiang2019}
F.~Jiang, H.~Zhao, J.~Du, Y.~Xue, and Y.~Peng, ``Outlier detection based on approximation accuracy entropy,'' \emph{Int. J. Mach. Learn. Cybern.}, vol.~10, no.~9, pp. 2483--2499, 2019.

\bibitem{F2015}
F.~Maciá-Pérez, J.~V. Berna-Martinez, A.~F. Oliva, and M.~A.~A. Ortega, ``Algorithm for the detection of outliers based on the theory of rough sets,'' \emph{Decis. Support Syst.}, vol.~75, pp. 63--75, 2015.

\bibitem{Hu2010}
Q.~Hu, D.~Yu, W.~Pedrycz, and D.~Chen, ``Kernelized fuzzy rough sets and their applications,'' \emph{IEEE Trans. Knowl. Data Eng.}, vol.~23, no.~11, pp. 1649--1667, 2010.

\bibitem{Xia2023a}
S.~Xia, S.~Wu, X.~Chen, G.~Wang, X.~Gao, Q.~Zhang, E.~Giem, and Z.~Chen, ``Grrs: Accurate and efficient neighborhood rough set for feature selection,'' \emph{IEEE Trans. Knowl. Data Eng.}, vol.~35, no.~9, pp. 9281--9294, 2023.

\bibitem{Chen2010}
Y.~Chen, D.~Miao, and H.~Zhang, ``Neighborhood outlier detection,'' \emph{Expert Syst. Appl.}, vol.~37, no.~12, pp. 8745--8749, 2010.

\bibitem{Yuan2018}
Z.~Yuan, X.~Zhang, and S.~Feng, ``Hybrid data-driven outlier detection based on neighborhood information entropy and its developmental measures,'' \emph{Expert Syst. Appl.}, vol. 112, pp. 243--257, 2018.

\bibitem{Yuan2022a}
Z.~Yuan, H.~Chen, T.~Li, X.~Zhang, and B.~Sang, ``Multigranulation relative entropy-based mixed attribute outlier detection in neighborhood systems,'' \emph{IEEE Trans. Syst. Man Cybern. Syst.}, vol.~52, no.~8, pp. 5175--5187, 2022.

\bibitem{Gao2023}
L.~Gao, M.~Cai, and Q.~Li, ``A relative granular ratio-based outlier detection method in heterogeneous data,'' \emph{Inf. Sci.}, vol. 622, pp. 710--731, 2023.

\bibitem{ZhangYM2023}
X.~Zhang, Z.~Yuan, and D.~Miao, ``Outlier detection using three-way neighborhood characteristic regions and corresponding fusion measurement,'' \emph{IEEE Trans. Knowl. Data Eng.}, vol. Early Access, Sep. 2023.

\bibitem{Yuan2021}
Z.~Yuan, H.~Chen, T.~Li, and J.~Liu, ``Fuzzy information entropy-based adaptive approach for hybrid feature outlier detection,'' \emph{Fuzzy Sets Syst.}, vol. 421, pp. 1--28, 2021.

\bibitem{Yuan2022}
Z.~Yuan, H.~Chen, T.~Li, and B.~Sang, ``A new entropy-based approach to outlier detection for multidimensional data,'' \emph{IEEE Trans. Fuzzy Syst.}, vol.~30, no.~9, pp. 3265--3276, 2022.

\bibitem{Yuan2023a}
Z.~Yuan, H.~Chen, C.~Luo, and D.~Peng, ``Mfgad: Multi-fuzzy granules anomaly detection,'' \emph{Inf. Fusion}, vol.~95, pp. 17--25, 2023.

\bibitem{Yuan2023b}
Z.~Yuan, B.~Chen, J.~Liu, H.~Chen, D.~Peng, and P.~Li, ``Anomaly detection based on weighted fuzzy-rough density,'' \emph{Appl. Soft. Comput.}, vol. 134, p. 109995, 2023.

\bibitem{Han2022}
S.~Han, X.~Hu, H.~Huang, M.~Jiang, and Y.~Zhao, ``Adbench: Anomaly detection benchmark,'' in \emph{Proc. Conf. on Neural Information Processing Systems (NeurIPS)}, 2022, pp. 32\,142--32\,159.

\bibitem{Tan2023}
X.~Tan, C.~Gao, J.~Zhou, and J.~Wen, ``Three-way decision-based co-detection for outliers,'' \emph{Int. J. Approx. Reasoning}, vol. 160, p. 108971, 2023.

\bibitem{Dubois1990}
D.~Dubois and H.~Prade, ``Rough fuzzy sets and fuzzy rough sets,'' \emph{Int. J. Gen. Syst.}, vol.~17, pp. 191--209, 1990.

\bibitem{Xia2022a}
S.~Xia, D.~Peng, D.~Meng, C.~Zhang, G.~Wang, E.~Giem, W.~Wei, and Z.~Chen, ``Ball $k$-means: Fast adaptive clustering with no bounds,'' \emph{IEEE Trans. Pattern Anal. Mach. Intell.}, vol.~44, no.~1, pp. 87--99, 2022.

\bibitem{YaoVP2013}
J.~Yao, A.~V. Vasilakos, and W.~Pedrycz, ``Granular computing: Perspectives and challenges,'' \emph{IEEE Trans. Cybern.}, vol.~43, no.~6, pp. 1977--1989, 2013.

\bibitem{Xia2019}
S.~Xia, G.~Wang, Z.~Chen, Y.~Duan, and Q.~Liu, ``Complete random forest based class noise filtering learning for improving the generalizability of classifiers,'' \emph{IEEE Trans. Knowl. Data Eng.}, vol.~31, no.~11, pp. 2063--2078, 2019.

\bibitem{Xie2023}
J.~Xie, W.~Kong, S.~Xia, G.~Wang, and X.~Gao, ``An efficient spectral clustering algorithm based on granular-ball,'' \emph{IEEE Trans. Knowl. Data Eng.}, vol.~35, no.~9, pp. 9743--9753, 2023.

\bibitem{Xia2022b}
S.~Xia, H.~Zhang, W.~Li, G.~Wang, E.~Giem, and Z.~Chen, ``Gbnrs: A novel rough set algorithm for fast adaptive attribute reduction in classification,'' \emph{IEEE Trans. Knowl. Data Eng.}, vol.~34, no.~3, pp. 1231--1242, 2022.

\bibitem{Xia2023}
S.~Xia, S.~Zheng, G.~Wang, X.~Gao, and B.~Wang, ``Granular ball sampling for noisy label classification or imbalanced classification,'' \emph{IEEE Trans. Neural Networks Learn. Syst.}, vol.~34, no.~4, pp. 2144--2155, 2023.

\bibitem{Zhang2017}
Q.~Zhang, S.~Yang, and G.~Wang, ``Measuring uncertainty of probabilistic rough set model from its three regions,'' \emph{IEEE Trans. Syst. Man Cybern. Syst.}, vol.~47, no.~12, pp. 3299--3309, 2017.

\bibitem{Platt1998}
J.~C. Platt, ``Sequential minimal optimization: A fast algorithm for training support vector machines,'' in \emph{Advances in Kernel Methods-Support Vector Learning}, 1998, pp. 1--21.

\bibitem{Campos2016}
G.~O. Campos, A.~Zimek, J.~Sander, R.~J. G.~B. Campello, B.~Micenkova, E.~Schubert, I.~Assent, and M.~E. Houle, ``On the evaluation of unsupervised outlier detection: Measures, datasets, and an empirical study,'' \emph{Data Min. Knowl. Discov.}, vol.~30, no. 9-10, p. 891–927, 2016.

\bibitem{Scholkopf2001}
B.~Schölkopf, J.~C. Platt, J.~C. Shawe-Taylor, A.~J. Smola, and R.~C. Williamson, ``Estimating the support of a high-dimensional distribution,'' \emph{Neural Comput.}, vol.~13, no.~7, p. 1443–1471, 2001.

\bibitem{Liu2008}
F.~T. Liu, K.~M. Ting, and Z.-H. Zhou, ``Isolation forest,'' in \emph{IEEE Int’l Conf. on Data Mining}, 2008, pp. 413--422.

\bibitem{Zhao2019}
Y.~Zhao, Z.~Nasrullah, M.~K. Hryniewicki, and Z.~Li, ``Lscp: Locally selective combination in parallel outlier ensembles,'' in \emph{Proc. SIAM Int’l Conf. on Data Mining}, 2019, pp. 585--593.

\bibitem{Pevny2016}
T.~Pevný, ``Loda: Lightweight on-line detector of anomalies,'' \emph{Mach. Learn.}, vol. 102, no.~2, pp. 275--304, 2016.

\bibitem{Demšar2006}
J.~Demšar, ``Statistical comparisons of classifiers over multiple data sets,'' \emph{J. Mach. Learn. Res.}, vol.~7, pp. 1--30, 2006.

\end{thebibliography}


\begin{IEEEbiography}[{\includegraphics[width=1in,height=1.25in,clip,keepaspectratio]{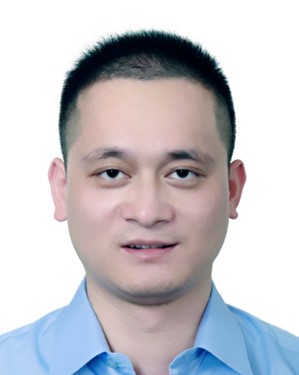}}]{Can Gao (Member, IEEE)}
received the Ph.D. degree in Pattern Recognition and Intelligent Systems from the Tongji University, Shanghai, China, in 2013. From 2010 to 2011, he was a Visiting Scholar with the University of Alberta, Edmonton, Canada. From 2015 to 2018, he was a Research Associate, a Post-Doctoral Fellow, and a Research Fellow with the Hong Kong Polytechnic University, Kowloon, Hong Kong. He is currently an Associate Professor with the College of Computer Science and Software Engineering, Shenzhen University, Shenzhen, China. He has authored or co-authored more than 80 academic papers. His research interests include anomaly detection, granular computing, and machine learning.
\end{IEEEbiography}

\begin{IEEEbiography}[{\includegraphics[width=1in,height=1.25in, clip,keepaspectratio]{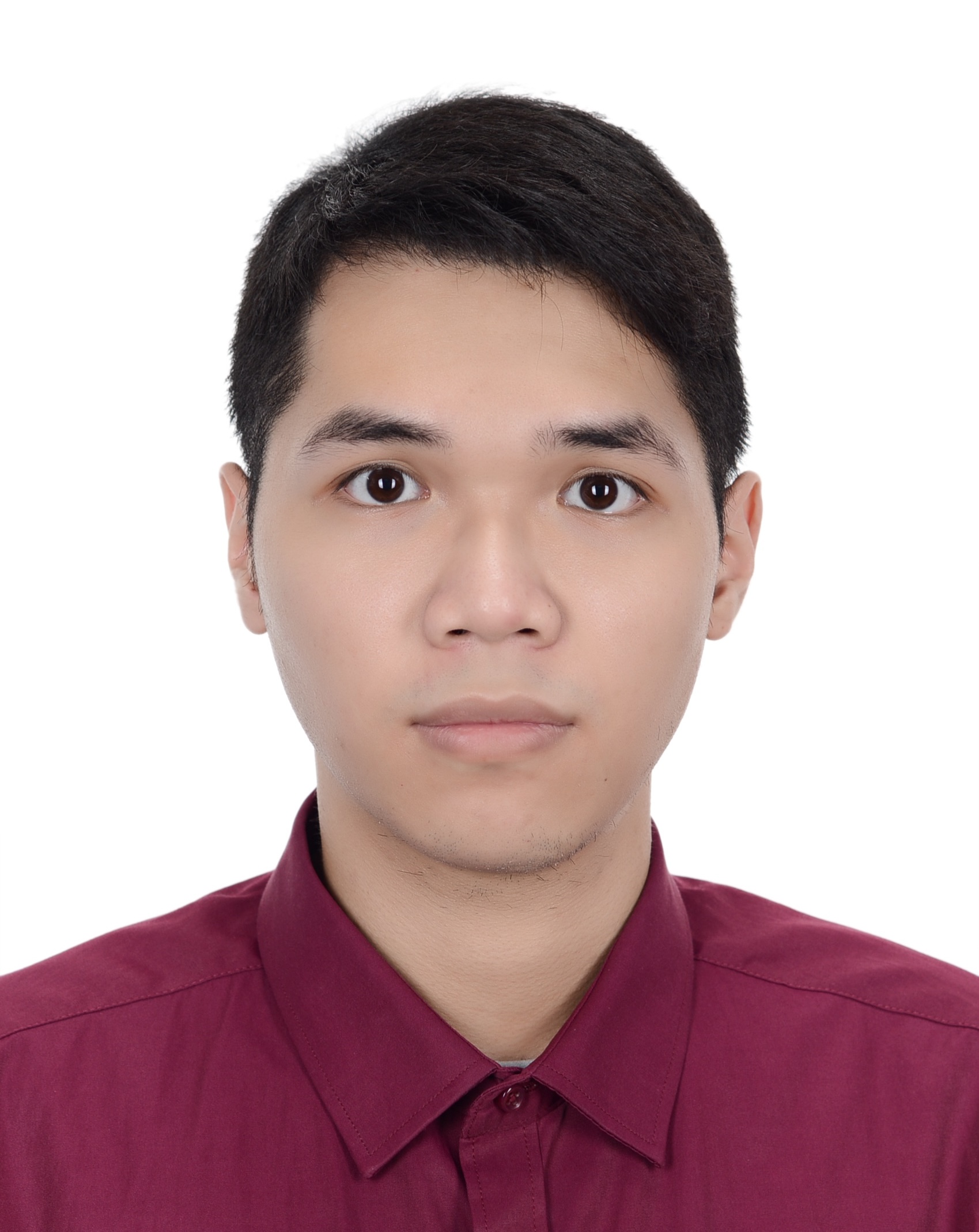}}]{Xiaofeng Tan} received the B.S. degree in Software Engineering from the Shenzhen University, Shenzhen, China. He is currently pursuing the M.E. degree with the School of Computer Science and Engineering, Southeast University, Nanjing, China. His research interests include anomaly detection and 3D computer vision.
\end{IEEEbiography}

\begin{IEEEbiography}[{\includegraphics[width=1in,height=1.25in,clip,keepaspectratio]{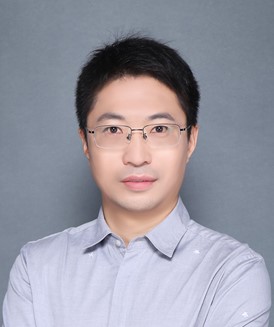}}]{Jie Zhou (Member, IEEE)}
received the Ph.D. degree in Pattern Recognition and Intelligent Systems from the Tongji University, Shanghai, China, in 2011. From 2010 to 2011, he was a Visiting Scholar with the University of Alberta, Edmonton, Canada. From 2017 to 2018, he was a Research Associate with the Hong Kong Polytechnic University, Kowloon, Hong Kong. He is currently an Associate Professor with the National Engineering Laboratory for Big Data System Computing Technology, Shenzhen University, Shenzhen, China. His current major research interests include uncertainty analysis, pattern recognition, data mining, and intelligent systems.
\end{IEEEbiography}

\begin{IEEEbiography}[{\includegraphics[width=1in,height=1.25in,clip,keepaspectratio]{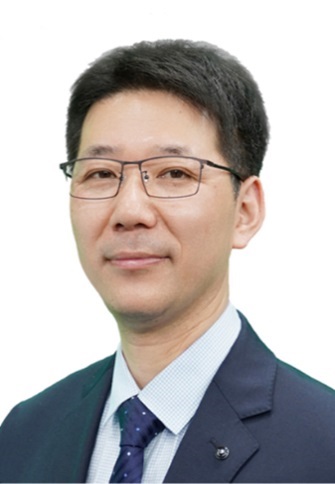}}]{Weiping Ding (M’16-SM’19)}
received the Ph.D. degree in Computer Science from the Nanjing University of Aeronautics and Astronautics, Nanjing, China, in 2013. From 2014 to 2015, he was a Postdoctoral Researcher at the Brain Research Center, National Chiao Tung University, Hsinchu, Taiwan, China. In 2016, he was a Visiting Scholar at the National University of Singapore, Singapore. From 2017 to 2018, he was a Visiting Professor at the University of Technology Sydney, Australia. His main research directions involve deep neural networks, granular data mining, and multimodal machine learning. He ranked within the top 2\% Ranking of Scientists in the World by Stanford University (2020-2024). He has published over 350 articles, including over 160 IEEE Transactions papers. His twenty authored/co-authored papers have been selected as ESI Highly Cited Papers. He has co-authored five books. He holds more 50 approved invention patents, including three U.S. patents and one Australian patent. He serves as an Associate Editor/Area Editor/Editorial Board member of more than 10 international prestigious journals, such as \textit{IEEE Transactions on Neural Networks and Learning Systems, IEEE Transactions on Fuzzy Systems, IEEE/CAA Journal of Automatica Sinica, IEEE Transactions on Emerging Topics in Computational Intelligence, IEEE Transactions on Intelligent Transportation Systems, IEEE Transactions on Intelligent Vehicles, IEEE Transactions on Artificial Intelligence, Information Fusion, Information Sciences, Neurocomputing, Applied Soft Computing, Engineering Applications of Artificial Intelligence, Swarm and Evolutionary Computation, et al}. He was/is the Leading Guest Editor of Special Issues in several prestigious journals, including \textit{IEEE Transactions on Evolutionary Computation, IEEE Transactions on Fuzzy Systems, Information Fusion, Information Sciences, et al}. Now he is the Co-Editor-in-Chief of three international journals: \textit{the Journal of Artificial Intelligence and Systems, Journal of Artificial Intelligence Advances, and Sustainable Machine Intelligence Journal}.
\end{IEEEbiography}

\begin{IEEEbiography}[{\includegraphics[width=1in,height=1.25in,clip,keepaspectratio]{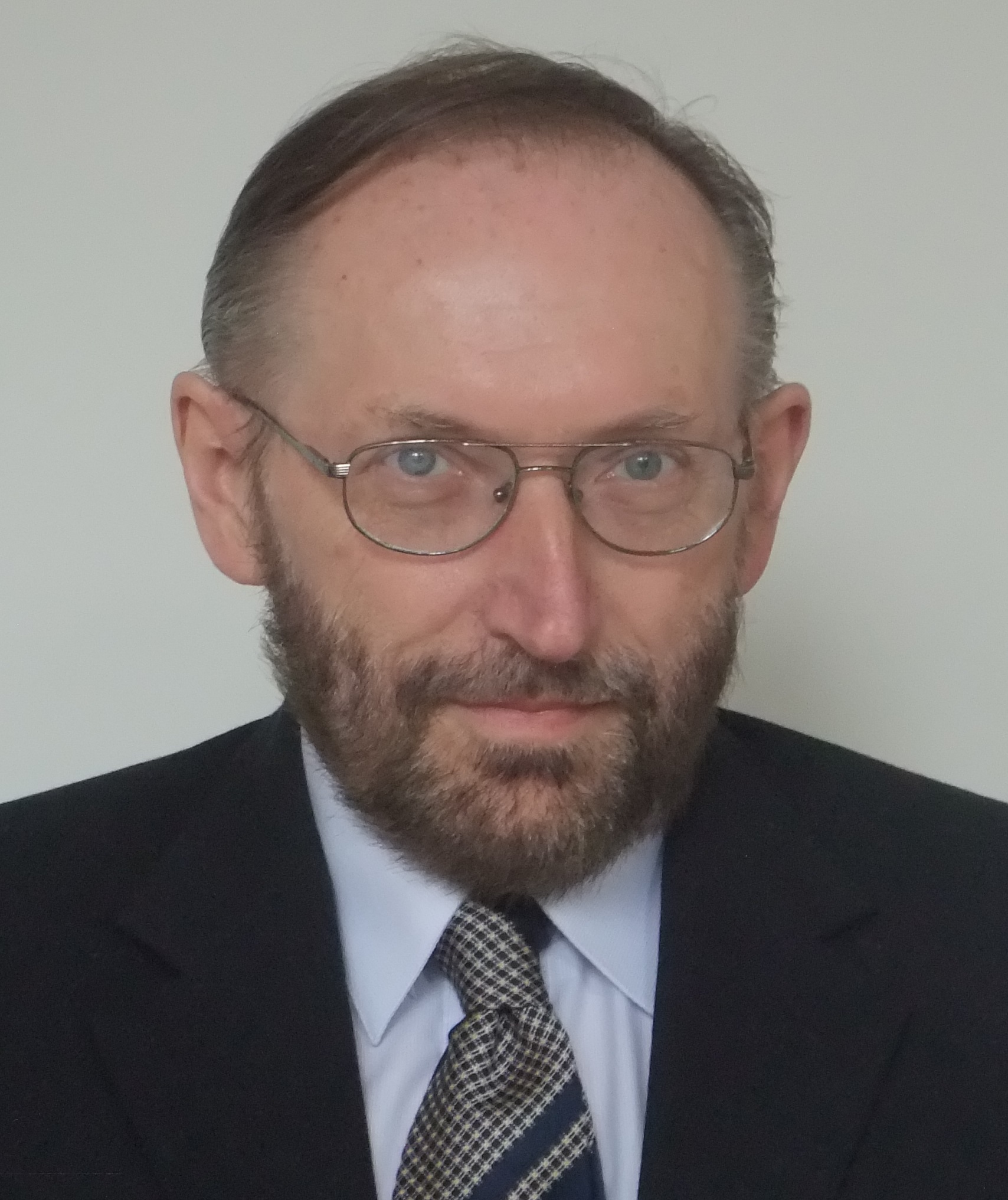}}]
{Witold Pedrycz (Life Fellow, IEEE)}
received the M.Sc. degree in Computer Science, the Ph.D. degree in Computer Engineering, and the D.Sci. degree in Systems Science from the Silesian University of Technology, Gliwice, Poland, in 1977, 1980, and 1984, respectively. He is a Professor and the Canada Research Chair of Computational Intelligence with the Department of Electrical and Computer Engineering, University of Alberta, Edmonton, AB, Canada. He is also with the System Research Institute, Polish Academy of Sciences, Warsaw, Poland, with the Department of Electrical and Computer Engineering, King Abdulaziz University, Jeddah, Saudi Arabia, and with the Department of Computer Engineering, Istinye University, Istanbul, Turkey. He has authored 15 research monographs covering various aspects of computational intelligence, data mining, and software engineering. His current research interests include computational intelligence, fuzzy modeling, granular computing, knowledge discovery and data mining, fuzzy control, pattern recognition, knowledge-based neural networks, relational computing, and software engineering.

Prof. Pedrycz was a recipient of the IEEE Canada Computer Engineering Medal, the Cajastur Prize for Soft Computing from the European Centre for Soft Computing, the Killam Prize, and the Fuzzy Pioneer Award from the IEEE Computational Intelligence Society. He is intensively involved in editorial activities. He is an Editor-in-Chief of \textit{Information Sciences} and \textit{WIREs Data Mining and Knowledge Discovery (Wiley)}. He currently serves as a member of a number of editorial boards of other international journals. He is a foreign member of the Polish Academy of Sciences and a Fellow of the Royal Society of Canada.
\end{IEEEbiography}

\end{document}